\documentclass[journal]{IEEEtran}

\usepackage{graphicx}
\usepackage{subfigure}
\usepackage{overpic}
\usepackage{amsmath}
\usepackage{flushend}
\usepackage{amssymb}
\usepackage{multirow}
\usepackage{makecell}
\usepackage{array}
\usepackage{stfloats}
\usepackage{algorithm}
\usepackage{algpseudocode}
\usepackage{amsmath}

\begin{document}

\title{Adaptive DropBlock Enhanced Generative Adversarial Networks for Hyperspectral Image Classification}

\author{Junjie Wang, Feng Gao, Junyu Dong, Qian Du

\thanks{This work was supported in part by the National Key
Research and Development Program of China under Grant 2018AAA0100602, in part by the National Natural Science Foundation of China under Grant U1706218, and in part by the Key Research and Development Program of Shandong Province under Grant 2019GHY112048. {\it (Corresponding author: Feng Gao.)}}
\thanks{J. Wang, F. Gao and J. Dong are with the Qingdao Key Laboratory of Mixed Reality and Virtual Ocean, School of Information Science and Engineering, Ocean University of China, Qingdao 266100, China.}
\thanks{Q. Du is with the Department of Electrical and Computer Engineering, Mississippi State University, Mississippi State, MS 39762 USA.}}

% The paper headers
\markboth{IEEE TRANSACTIONS ON GEOSCIENCE AND REMOTE SENSING}%
{Shell}

\maketitle
\begin{abstract}

In recent years, hyperspectral image (HSI) classification based on generative adversarial networks (GAN) has achieved great progress. GAN-based classification methods can mitigate the limited training sample dilemma to some extent. However, several studies have pointed out that existing GAN-based HSI classification methods are heavily affected by the imbalanced training data problem. The discriminator in GAN always contradicts itself and tries to associate fake labels to the minority-class samples, and thus impair the classification performance. Another critical issue is the mode collapse in GAN-based methods. The generator is only capable of producing samples within a narrow scope of the data space, which severely hinders the advancement of GAN-based HSI classification methods. In this paper, we proposed an \underline{A}daptive \underline{D}ropBlock-enhanced \underline{G}enerative \underline{A}dversarial \underline{N}etworks (ADGAN) for HSI classification. First, to solve the imbalanced training data problem, we adjust the discriminator to be a single classifier, and it will not contradict itself. Second, an adaptive DropBlock (AdapDrop) is proposed as a regularization method employed in the generator and discriminator to alleviate the mode collapse issue. The AdapDrop generated drop masks with adaptive shapes instead of a fixed size region, and it alleviates the limitations of DropBlock in dealing with ground objects with various shapes. Experimental results on three HSI datasets demonstrated that the proposed ADGAN achieved superior performance over state-of-the-art GAN-based methods. Our codes are available at \verb'https://github.com/summitgao/HC_ADGAN'

\end{abstract}


\IEEEpeerreviewmaketitle

\section{Introduction}

\IEEEPARstart{B}{enifiting} from the advancement of earth observation programs, hyperspectral sensors have received great attention over the past few years. A great number of hyperspectral images (HSI) captured by spaceborne or airborne sensors are available \cite{Tu19_tgrs}. These images have high spectral resolutions and abundant spatial information, which brings opportunities to a wide variety of applications, such as urban development \cite{Benediktsson05_tgrs}, land cover change monitoring \cite{Wang19_tgrs}, environmental pollution monitoring \cite{Arellano15_ep} and resource management \cite{Wang16_tgrs}. Among these applications, classification has become one of the most critical topics in the hyperspectral application community.

Hyperspectral images classification aims to assign a distinct label to each pixel vector so that it is well defined by a given class. Plenty of methods have been proposed to solve the problem. In the early days of HSI classification, researchers mainly focused on spectral information \cite{Ham05_tgrs} \cite{Ratle10_tgrs} \cite{Fauvel06_icassp}. However, the same object in different locations may exhibit different spectral features, while different objects may emerge with similar spectral features \cite{Zhang19_tgrs}. It is commonly difficult to classify such objects by using spectral features alone. To solve the problem, many studies combined spectral features with spatial features to establish spectral-spatial models for HSI classification. Benediktsson \emph{et al.} \cite{Benediktsson05_tgrs} proposed a classification method based on mathematical morphology profiles, which uses both the spatial and spectral features for classification. Fauvel \emph{et al.} \cite{Fauvel09_tgrs} established a framework that fused the morphological information and the original hyperspectral image. Li \emph{et al.} \cite{Li15_tgrs1} presented a classification framework that integrates the local binary patters (LBP), global Gabor features and spectral features. In \cite{Zou15_grsl}, spatial-spectral information was transposed into a sparse model for classification. Pan \emph{et al.} \cite{Pan17_tgrs} developed the hierarchical guidance filtering to obtain a set of spectral-spatial features from different scales, and then an ensemble model is established to utilize these features simultaneously. Besides these techniques, morphological kernel \cite{Fang15_tgrs} \cite{Fauvel12_pr}, edge-preserving filter \cite{Kang14_tgrs}, extinction profile \cite{Fang18_tgrs} and superpixel segmentation \cite{Li16_tgrs} are also employed to explore spectral and spatial information for HSI classification. The combination of spectral and spatial information improves the classification performance \cite{Zhong14_jstars} \cite{Tang15_tgrs} \cite{Li15_tgrs2} \cite{Chen14_jstars1}. Although these techniques have achieved excellent performance, they are mainly based on hand-crafted descriptors. However, most hand-crafted descriptors heavily depend on prior knowledge to obtain optimal parameters, which limits the applicability of these methods in various scenarios. Robust feature extraction is widely acknowledged as a critical step in HSI classification.

Deep learning has become the most impactful developments in artificial intelligence and big data analysis over the past few years. It has been demonstrated that deep models are capable of extracting the invariant and discriminant features efficiently in computer vision and natural language processing tasks \cite{Krizhevsky12_nips} \cite{Girshick14_cvpr} \cite{Hinton12_spm} \cite{Yao13_interspeech}. Inspired by these flourishing tehniques, deep models have been designed to classify HSIs. Chen \emph{et al.} \cite{Chen14_jstars2} first presented a deep learning-based HSI classification method, and employed a stacked autoencoder (SAE) as a classifier. In \cite{Chen15_jstars}, the deep belief network (DBN) is introduced for spectral-spatial information exploration. Pan \emph{et al.} \cite{Pan17_jstars} proposed a vertex component analysis network (VCANet), which takes the physical characteristics of HSIs into account. VCANet is capable to exploit discriminative features when training samples are limited.

Recently, convolutional neural networks (CNNs) have been widely used in HSI classification. CNNs make use of regional connections to extract contextual features, and have shown outstanding classification performance. In \cite{Zhang17_rsl}, spectral features are extracted via one-dimensional CNN, and spatial features are exploited via two-dimensional CNN. Then spectral and spatial features are combined for classification. Chen \emph{et al.} \cite{Chen16_tgrs} developed a 3D-CNN HSI classification model, in which $L_2$ regularization is used in the training procedure to mitigate the overfitting problem. Zhong \emph{et al.} \cite{Zhong18_tgrs} presented an end-to-end CNNs that take the 3D cube as input data. Residual learning is introduced to solve the exploding gradient problem. In \cite{Chen17_grsl}, Gabor filters are combined with convolutional filters to alleviate the overfitting problem in CNN training. Inspired by the inception module \cite{Szegedy15_cvpr}, Gong \emph{et al.} \cite{Gong19_tgrs} proposed a CNNs with multiscale convolution. The multiscale filter banks enrich the representation power of the deep model. Ma \emph{et al.} \cite{Ma18_tgrs} designed an end-to-end deconvolution network with skip architecture for spatial and spectral feature extraction. The network is capable to recover the lost information in the pooling operation of the CNN via unpooling and deconvolution layers.

CNN-based methods have achieved tremendous progress in HSI classification. However, the performance of these techniques heavily depends on the number of training samples. Commonly, it is a challenging task to collect lots of training samples from HSIs. This problem can be alleviated by data augmentation. Cropping, horizontal flipping, generative model are typical data augmentation techniques. Recently, the generative model has drawn a lot of attention, since it is able to generate high-quality samples to alleviate the overfitting problem. Goodfellow \emph{et al.} \cite{Goodfellow14_nips} designed the generative adversarial network (GAN), which is comprised of a generator $G$ and a discriminator $D$. The generator $G$ captures the data distribution while $D$ judges that whether a sample comes from $G$ or from the training data. The generator $G$ can be considered as a regularization method that can effectively alleviate the overfitting problem to a great extent.

Researchers have made efforts to design GAN-based models to alleviate the limited high-quality sample problem. Zhan \emph{et al.} \cite{Zhan18_grsl} proposed a semisupervised framework based on 1D-GAN. After that, Zhu \emph{et al.} \cite{Zhu18_tgrs} proposed a 3D-GAN for HSI classification. The spatial information is taken into consideration, and a softmax classifier is employed in the discriminator $D$ to auxiliary classification. Feng \emph{et al.} \cite{Feng19_tgrs} proposed a multiclass  GAN for HSI classification. Two generators are designed in multiclass GAN to generate hyperspectral image patches, and a discriminator is devised to output multiclass probabilities. Zhong \emph{et al.} \cite{Zhong19_tc} integrated GAN and conditional random field (CRF) together, where dense CRFs impose graph constraints on the discriminators of GAN to refine the classification results.

Although these GAN-based models have achieved satisfying performance over their contemporary baselines, there still exist two drawbacks over HSI classification, which are urgently needed to be solved.

The first challenge is \emph{imbalanced training data}. The accuracy of classification is likely to deteriorate when available training samples are not uniformly distributed among different classes. However, the imbalanced training data problem is fundamental in HSIs, since objects with different sizes present in a typical scene \cite{Li18_tgrs}. In Zhu's work \cite{Zhu18_tgrs}, auxiliary classifier GAN (ACGAN) \cite{Odena16_icml} is employed for HSI classification. In ACGAN, the discriminator has two outputs: one to discriminate real and fake samples, and the other to classify samples. It seems that ACGAN is capable of producing samples of a specific class. In practice, it is observed that two loss functions of the discriminator turnout to be flawed when generating the minority-class samples. The reason for this phenomenon is that when minority-class samples are passed to the discriminator, they are likely to be assigned the fake label. Therefore, the discriminator intends to associate fake label to the minority-class samples. At this point, the generators produce samples that look real but not represent the minority class. The quality of generated samples is deteriorated, and hence the classification performance is impaired.

Another critical issue is \emph{mode collapse}. The generator fools the discriminator by only producing data from the same data mode \cite{Chavdarova18_cvpr}. It leads to a weak generator that can generate samples within a narrow scope of the data space. Therefore, the generated samples are too similar for the model to learn the true data distribution, and the model can hardly learn the full data distribution. The model collapse can be considered as a consequence of overfitting to the feedback of the discriminator. In a disparate line of work, DropBlock \cite{Ghisasi18_nips} was designed in CNNs to alleviate overfitting. In DropBlock, features in a square mask from one feature map are dropped together during training. It is demonstrated that DropBlock can learn more spatially distributed representations. However, when dealing with objects with various shapes, the fixed square masks are inflexible. We argue that if irregularly shaped masks are taken into account, the mode collapse problem can be alleviated.

To tackle the aforementioned limitations of GAN-based classification methods, we established an \underline{A}daptive \underline{D}ropblock-enhanced \underline{G}enerative \underline{A}dversarial \underline{N}etworks (ADGAN) for HSI classification. On the one hand, considering the contradiction between the loss functions of the discriminator in ACGAN, the discriminator in ADGAN is adjusted to be one single output that returns either the specific class label or the fake label. The generator is trained to avoid the fake label and match the desired class labels. Since the discriminator is now defined as one single objective, it will not contradict itself. On the other hand, we propose adaptive DropBlock (AdapDrop) as a regularization method used in the generator and the discriminator. Instead of dropping a fixed size region, the AdapDrop generated drop masks with adaptive shapes, relaxing the limitations of DropBlock in dealing with objects with various shapes.

To validate the effectiveness of the proposed method, extensive experiments are conducted on three datasets. Experimental results demonstrate that the proposed ADGAN yields better performance than state-of-the-art GAN-based methods. In summary, the our contributions are threefold:

\begin{itemize}
  \item We develop a novel GAN-based HSI classification model that contains a single output in discriminator. The contradictions in ACGAN when dealing with minority-class samples are mitigated.

  \item For the purpose of alleviating the mode collapse problem, we propose the AdapDrop for regularization. The AdapDrop generates masks with adaptive shapes, which can boost the classification performance.

  \item We conducted extensive experiments on three well-known hyperspectral datasets under the condition of limited training samples to validate the effectiveness of the proposed method. The experimental results achieve competitive results compared with other state-of-the-art classification methods.

\end{itemize}

The rest of this paper is organized as follows. In Section II, the basic concepts of ACGAN and DropBlock are briefly reviewed. The scheme of the proposed method and its components are introduced in Section III. Experimental results and analysis are presented in Section IV. Finally, conclusions are drawn in Section V.

\section{Background}

\subsection{Generative Adversarial Networks}

In recent years, GANs provide a solution to estimate input data distribution, and correspondingly generate synthetic samples \cite{Goodfellow14_nips}. GAN is comprised of two parts, the generator $G$ and the discriminator $D$. The generator $G$ attempts to learn the distribution of real data, and generate data that subjects to this distribution. The discriminator $D$ judges whether the input is real or fake.The generator $G$ takes a random noise vector $z$ as input, and outputs an image $X_\textrm{fake} = G(z)$. The discriminator $D$ receives a real image or a synthesized image from $G$ as inputs, and outputs a probability distribution $P(S|X)=D(X)$. The discriminator $D$ is trained to maximize the log-likelihood it assigns to the correct source as follows:

\begin{equation}
\begin{split}
L_D=E[\log P(S=\textrm{real}|X_\textrm{real})] +
 ~~~~~~~~~~~~~~~~~~~~ \\
 E[\log P(S=\textrm{fake}|X_\textrm{fake})].
\end{split}
\end{equation}
The generator $G$ is trained to minimize the following likelihood:
\begin{equation}
L_G=E[\log P(S=\textrm{fake}|X_\textrm{fake})].
\end{equation}

When training GAN, an alternating optimization technique is employed. Specifically, $D$ is optimized by maximizing $L_D$ with $G$ fixed in one iteration. After that, $G$ is optimized with minimizing $L_G$ with $D$ fixed. In such adversarial training, the discriminator $D$ and the generator $G$ promote each other. After many iterations, $G$ captures the distribution of real data. At the same time, the capability of $D$ to distinguish real data and fake data is enhanced.

\subsection{Auxiliary Classifier GANs}

In naive GAN, the discriminator only judges whether the input samples are real or fake. Therefore, they are not suitable for multiclass image classification. To tackle the limitations of naive GAN, Odena \emph{et al.} \cite{Odena16_icml} proposed auxiliary classifier GAN (ACGAN). In ACGAN, the discriminator $D$ is a softmax classifier that can output multiclass label probabilities.

\begin{figure}[ht]
\centering
\begin{center}
\includegraphics [width=3.3in]{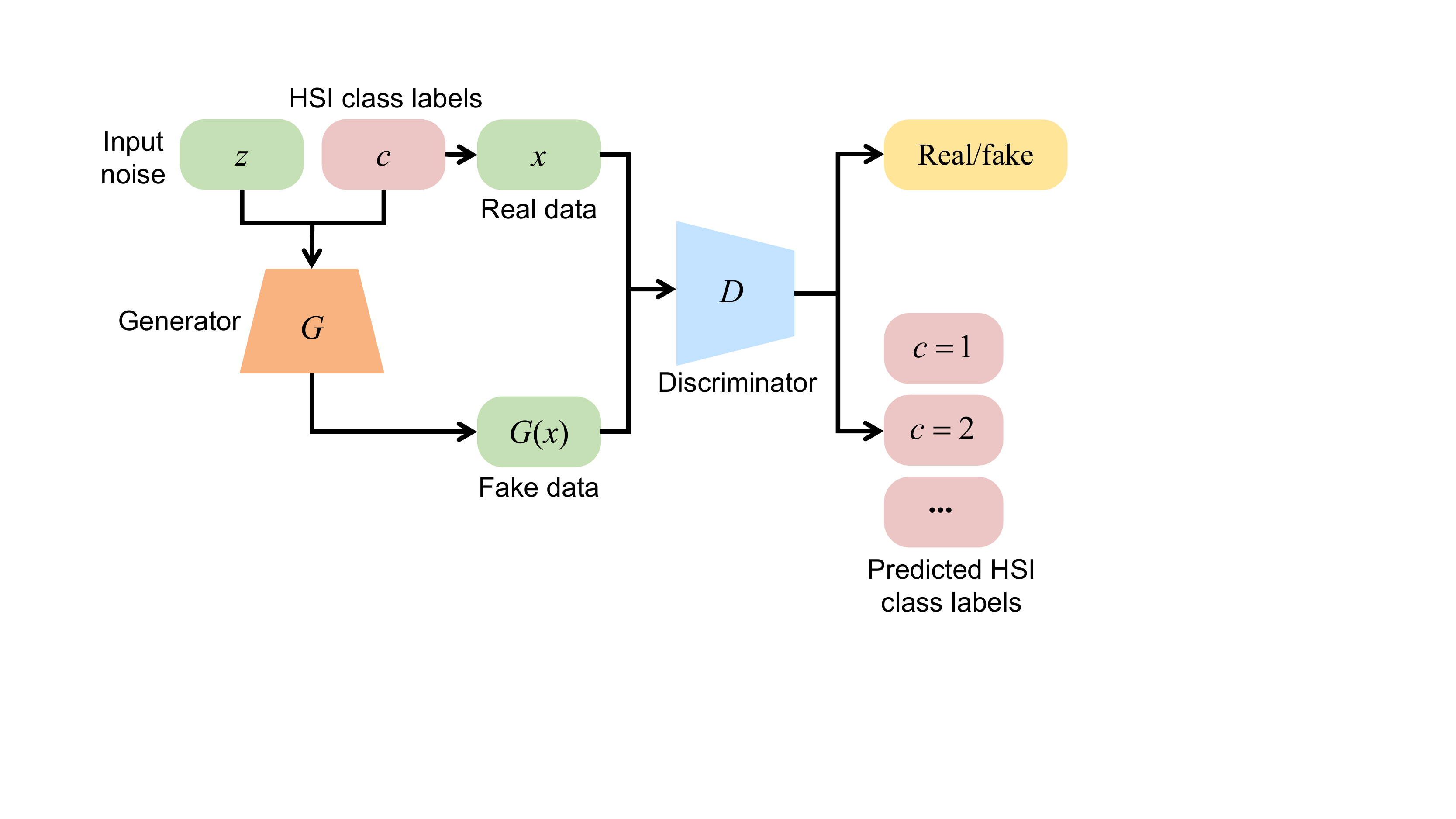}
\caption{Architecture of ACGAN employed in \cite{Zhu18_tgrs} for HSI classification}
\label{fig_acgan}
\end{center}
\end{figure}

The architecture of ACGAN employed in \cite{Zhu18_tgrs} is illustrated in Fig. \ref{fig_acgan}. The generator $G$ accepts the class label $c$ as input. The real data with the corresponding label and the data generated by $G$ are regarded as the input of $D$. The discriminator $D$ has two outputs: one to discriminate the real and fake data, and the other to classify input in terms of its class $c$. The loss function of ACGAN is comprised of two parts: the log-likelihood of the right source of input $L_S$ and the log-likelihood of the right class labels $L_C$. The $L_S$ and $L_C$ are computed as:

\begin{equation}
\begin{split}
L_{S}=E[\log P(S=\textrm{real}|X_\textrm{real})]+
~~~~~~~ \\
E[\log P(S=\textrm{fake}|X_\textrm{fake})],
\end{split}
\end{equation}

\begin{equation}
\begin{split}
L_{C} = E[\log P(C=c|X_\textrm{real})]+
~~~~~~~~ \\
E[\log P(C=c|X_\textrm{fake})].
\end{split}
\end{equation}

During training, the generator $G$ is optimized to maximize $L_C-L_S$, and the discriminator $D$ is optimized to maximize $L_S+L_C$. Therefore, the generator $G$ can be conditioned to draw a sample of the desired class.

\subsection{DropBlock}

Most current deep models are inclined to suffer from over-parameterization, and therefore give rise to overfitting problem. In this regard, regularization methods are harnessed to mitigate this issue. To date, dropout \cite{Srivastava14_jmlr} is a widely used regularization method and has been proved to be rather effective for fully connected layers. However,  features in convolutional layers are highly spatially correlated. Dropout becomes less effective since it does not take image spatial information into account.

Recently, Ghisasi \emph{et al.} \cite{Ghisasi18_nips} proposed DropBlock, which is particularly effective to regularize the convolutional layers. Rather than dropping out random units, DropBlock drops contiguous regions from a feature map in convolutional layer. It can be considered as a form of structured dropout. It is demonstrated that employing DropBlock in convolutional layers and skip connections effectively improve the classification performance.

\section{Methodology}

\begin{figure*}[ht]
\begin{center}
\includegraphics [width=5.5in]{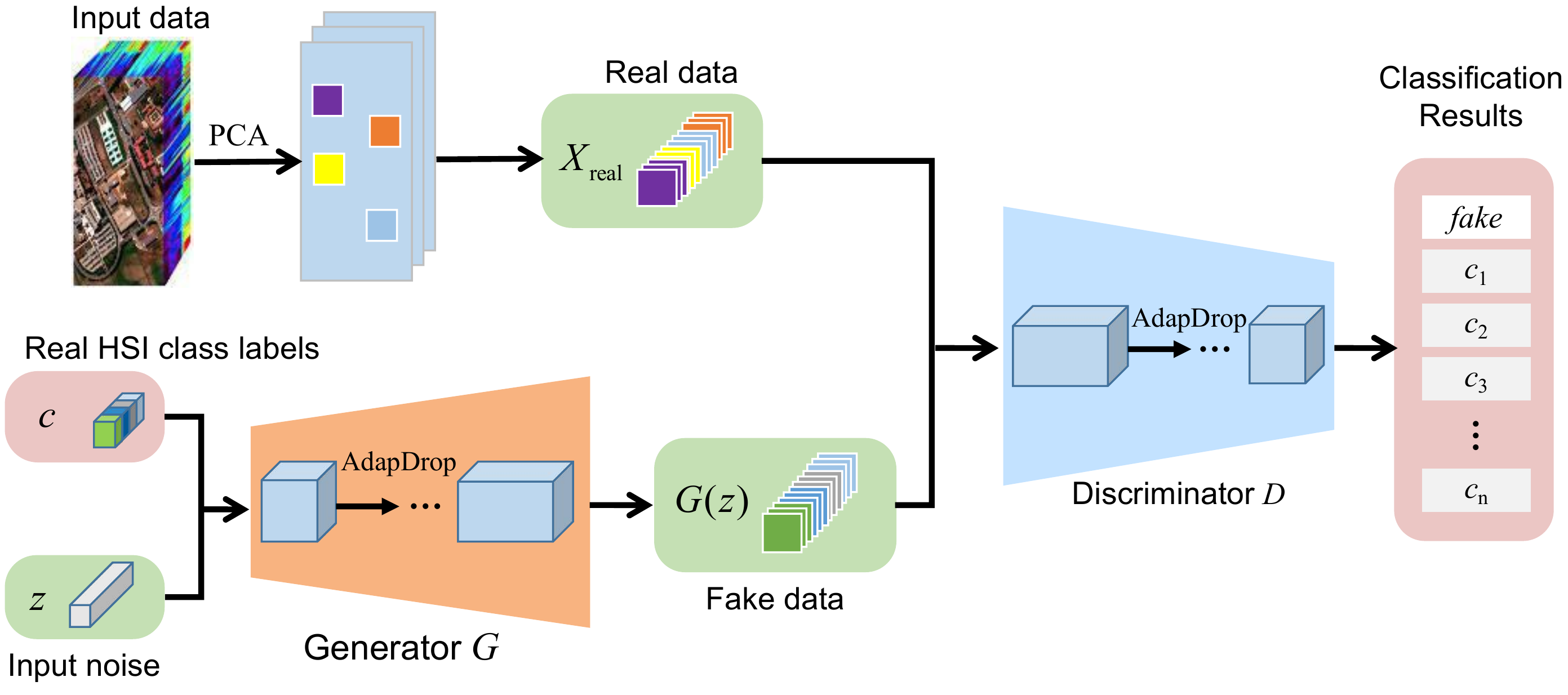}
\caption{Framework of ADGAN for HSI classification.}
\label{fig_framework}
\end{center}
\end{figure*}

\subsection{Framework of the Proposed ADGAN}

The framework of the proposed ADGAN is illustrated in Fig. \ref{fig_framework}. The input hyperspectral image contains hundreds of bands, and there is a lot of redundancy among these spectral bands. It is rather difficult to obtian a robust generator $G$, since the generator can hardly imitate the real data when high redundancy exists. Therefore, the number of spectral bands of the input HSI is reduced to three components by PCA \cite{Licciardi11_grsl}.  The spectral information can be condensed to a suitable scale by PCA. This operation is a non-trivial step since PCA can not only dramatically reduce the computational complexity, but also contribute to training a robust generator $G$.

From Fig. \ref{fig_framework}, it can be observed that the input of the generator $G$ include both noise $z$ and class labels $c$. The discriminator $D$ receives the image patches $X_\textrm{real}$ with labels $c$ and some fake patches $X_\textrm{fake}=G(z, c)$. It should be noted that in the proposed ADGAN, the discriminator $D$ has only one single output that returns either a specific class $c$ or the $fake$ label. Then, the generator $G$ is trained to generate image patches that match the desired class label. To this end, the discriminator $D$ is trained to maximize the log-likelihood as follows:
\begin{equation}
\begin{split}
L_D=E [ \log P( C=c|X_{\mathrm{real}})] +
~~~~~~~~~~~~~~~~~~~~~~ \\
E[ \log P( C=fake|X_{\mathrm{fake}} ) ].
\end{split}
\end{equation}
The generator $G$ is trained to maximize the log-likelihood as follows:
\begin{equation}
L_G=E[\log P(C=c|X_\textrm{fake})].
\end{equation}

The first term of $L_D$ encourages the discriminator $D$ to assign true label for real sample, and the second term expects to assign $fake$ label to the generated samples. On the contrary, the generator $G$ expects to draw a sample of the desired class. By adversarial learning, the generator $G$ captures the real data distribution of the desired class.

As mentioned before, the discriminator in Zhu's work \cite{Zhu18_tgrs} has two outputs, one to discriminate the real and fake data, and the other to classify the input in term of its class $c$. In the training phase, the generator aims to draw images belonging to class $c$. Therefore, the parameters of generator are optimized to maximize the superposition of two components. The first is the log-likelihood of generating an image that the discriminator considers real. The second component is the log-likelihood of generating an image that the discriminator considers it to be class $c$. However, there exists a contradiction between two components when dealing with the minority-class. Specifically, when a generated minority-class image is fed into the discriminator, it is likely to be judged as a fake image since the minority-class images are scarce in the training set. To optimize its loss function, the discriminator prefers to associate $fake$ label to the minority-class images. Then, the two components of generator will contradict each other, and two components can hardly be optimized at once. This phenomenon deteriorates the quality of generated images, which severely limits the performance of GAN-based approaches for HSI classification. To solve this problem, we proposed the ADGAN that alleviates the contradiction in ACGAN, and achieves the balance in training samples to some extent.

In the proposed ADGAN, the discriminator $D$ has one output that returns either a specific class $c$ or the $fake$ label, as shown in Fig. \ref{fig_framework}. The discriminator $D$ is trained to associate the real samples with their class label $c$. Meanwhile, $D$ also tries to associate the samples generated by $G$ with the $fake$ label. On the contrary, the generator $G$ is trained to avoid the $fake$ label and match the generated samples to the desired class. By doing so, the balance of training samples can be balanced to some extent. Besides, since the discriminator in ADGAN is now defined as one single objective rather than a combination of two objectives, it will not contradict itself.

In ADGAN, the network extracts the spectral and spatial feature simultaneously. The input hyperspectral data is condensed by PCA, and only three components are reserved. Through PCA, an optimal representation of the input HSI is achieved, and the computational burden is also dramatically reduced. From Fig. \ref{fig_framework}, we can observe that the generator $G$ accepts random Gaussian noise $z$ as input. The noise $z$ is transformed to the same size as the real input data with three bands in the spectral domain. After that, the discriminator $D$ accepts the generated fake samples togehter with the real samples as input. The output of $D$ indicates the probability that the input sample belongs to class $c$ or fake.  After many iterations, both $G$ and $D$ achieve optimized results. Specifically, $G$ can generate fake data that subject to the distribution of real data, while $D$ can hardly discriminate it. The competition between both networks can promote the HSI classification performance. The key idea of ADGAN lies in restoring the balance of dataset, and through the design of novel adversarial objective functions, the contradiction in ACGAN can be alleviated.

\begin{figure}[ht]
\begin{center}
\includegraphics [width=2.5in]{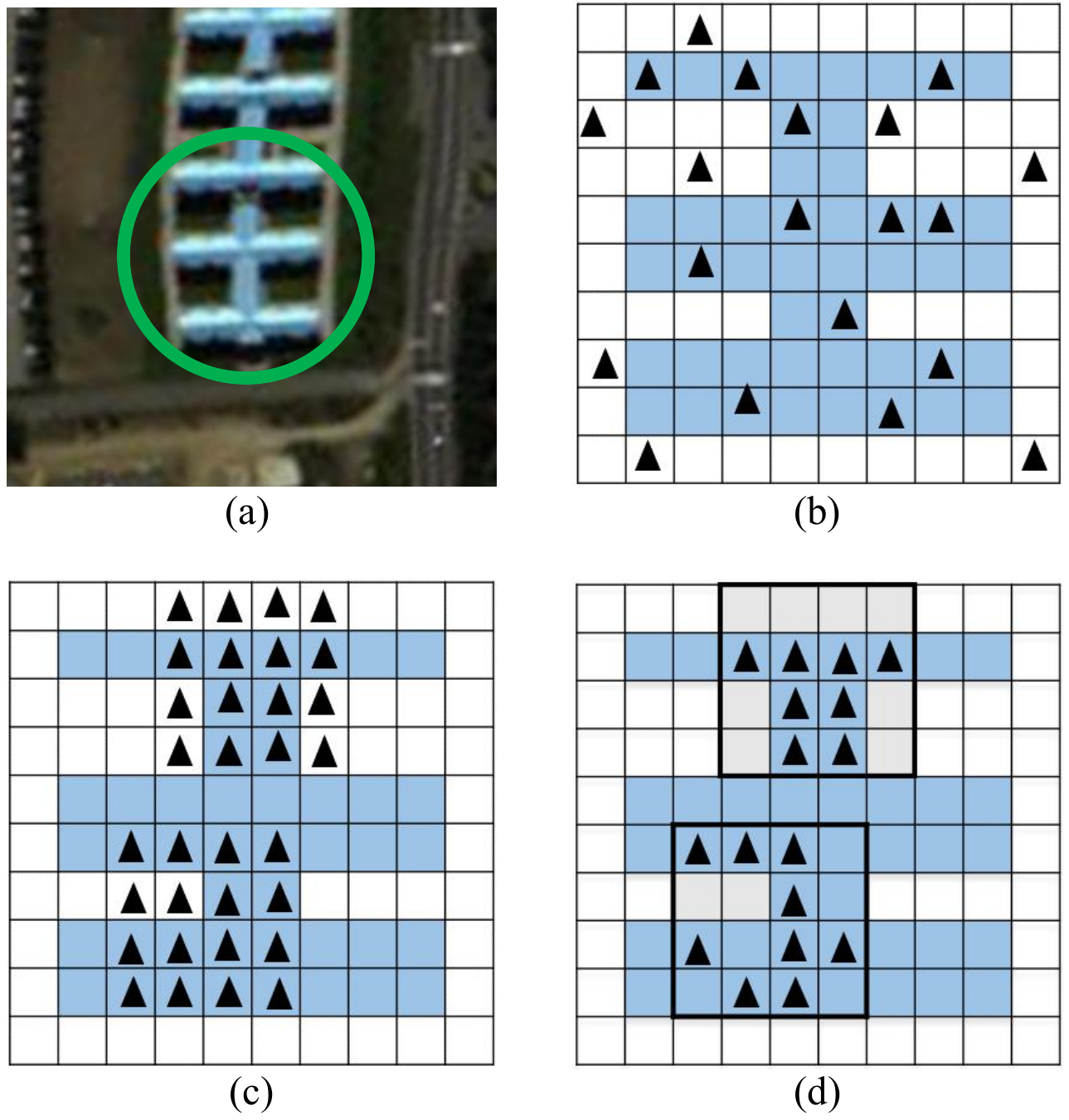}
\caption{Illustration of Dropout, DropBlock and AdapDrop. (a) input image. (b) Dropout. (c) DropBlock. (d) The proposed AdapDrop. The regions with meaningful information are marked by blue squares, and dropped operations are marked by black triangles. Dropout can hardly utilize spatial information. DropBlock takes advantage of the spatial information, but it breaks the meaningful semantic information for learning. The proposed AdapDrop drops pixels based on the activations of the input, and improves the model's capability of focusing on non-trival spatial information.}
\label{fig_adapdrop}
\end{center}
\end{figure}

\subsection{Structured Dropout with Attention Mechanism: Adaptive DropBlock}

In the proposed ADGAN, $G$ and $D$ are both in the form of convolutional networks. Pooling layers are replaced with strided convolutions. In addition, batch normalization is employed in both the $G$ and $D$. To further improve the classification performance, we proposed Adaptive DropBlock as the regularization method to enhance the spatially correlated feature representations.

Deep neural networks generally suffer from over-parameterization, and thus give rise to the overfitting problem. Regularization methods, such as batch normalization and dropout, are harnessed to mitigate the problem. In this paper, the proposed ADGAN is comprised of two convolutional networks, which makes it more complex to regularize.

Many regularization methods have been proposed to alleviate the overfitting problem, such as DropBlock \cite{Ghisasi18_nips}, DropPath \cite{Barret18_cvpr}, and DropConnect \cite{Wan13_icml}. DropBlock is one of the most commonly used regularization methods in CNNs. It operates on the feature maps, and then the units in a random contiguous region of feature maps are dropped together. As DropBlock discards features in a correlated area, the remaining semantic information will be used to fit the training data. However, DropBlock backfires when it drops overmuch information in the feature map, since it drops the whole blocks with fixed shape, which may contain essential features for training, as illustrated in Fig. \ref{fig_adapdrop}.

To mitigate the drawback of DropBlock, we propose \underline{Adap}tive \underline{Drop}Block (AdapDrop), which is a structured regularization method with attention mechanism. AdapDrop first randomly selects some blocks in the feature map. Then it produces adaptive mask with irregular shapes in the selected blocks by dropping the top $k$th percentile elements. The top $k$th percentile elements are chosen according to the values in the feature map. Due to the continuity of image pixel values, the neighboring pixels in the feature map have similar values. Hence, when we drop the top $k$th percentile elements, it is usually expressed as an irregular shape according to the spatial characteristics of the target object, such as the roof of a building, lawn garden, etc. Therefore, the AdapDrop effectively remove maximally activated regions and encourage the network to consider less prominent features.

\begin{figure}[ht]
\begin{center}
\includegraphics [width=3.3in]{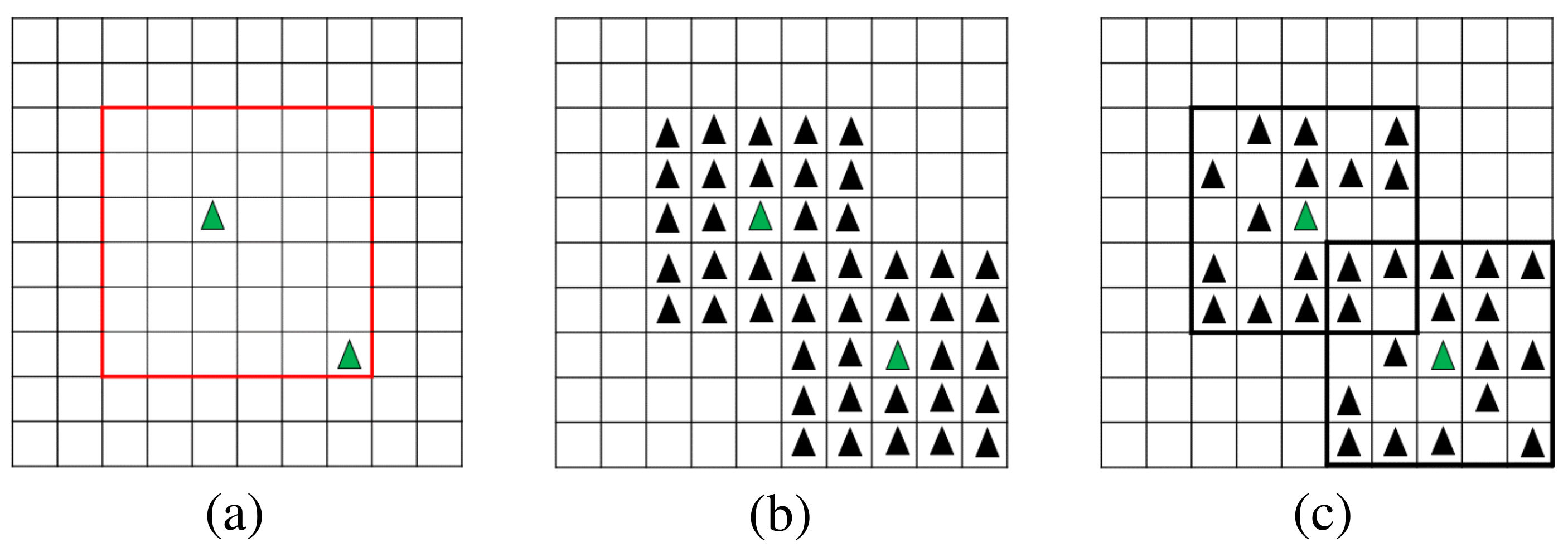}
\caption{Illustration of the proposed AdapDrop. (a) Similar as DropBlock, two elements are sampled from each feature map. (b) Every element is expanded to a $b\_size\times b\_size$ block. (c) In each block, the largest $k$th percentile elements are dropped.}
\label{fig_adapdrop_frame}
\end{center}
\end{figure}

The three phases of AdapDrop is shown in Fig. \ref{fig_adapdrop_frame}. An adaptive mask is applied to the feature map and scale the output. The AdapDrop algorithm is shown in Algorithm 1. It has three parameters, which are $b\_size$, $k$, and $\gamma$. $b\_size$ is the size of the mask block, and $k$ denotes the top $k$th percentile elements in the mask block will be dropped. $\gamma$ controls the number of features to drop, and the computation of $\gamma$ can be found in \cite{Ghisasi18_nips}. The current feature map $A^{(n)}$ is first normalized, and a new feature map $A'^{(n)}$ is generated. A set of pixels are sampled with the Bernoulli distribution. For each position $M_{i,j}$, create a spatial square block centered at $M_{i,j}$. The size of the block is $b\_size \times b\_size$. In each block, the top $k$th percentile elements are set to be zero, and the rest elements are set to be one. Next, apply the mask and scale the output as:
\begin{equation}
    A^{(n+1)} = A'^{(n)}\times
    \texttt{count}(M)/\texttt{count\_ones}(M)
\end{equation}
The $\texttt{count}(M)$ denotes the number of elements in the masks, and $\texttt{count\_ones}(M)$ denotes the number of one in the masks.

\begin{algorithm}[htb]
  \caption{The workflow of Adaptive DropBlock}
  \begin{algorithmic}[1]
    \Require
      Feature of the current layer $A^{(n)}$, $b\_size$, $\gamma$, $k$
    \Ensure
      Feature of the next layer $A^{(n+1)}$

    \State Normalize the feature:
      $$A^{'(n)}=\frac{A^{(n)}-A^{(n)}_{\min}}
      {A^{(n)}_{\max}-A^{(n)}_{\min}}$$
    \State Randomly sample some pixels: $M_{i,j}
    \thicksim \textrm{Bernoulli}(\gamma)$
    \State For each $M_{i,j}$, create a block centered at $M_{i,j}$. The size of the block is $b\_size \times b\_size$. In each block, set the top $k$th percentile elements to be zero, and set the rest elements to be one.
    \State Apply the mask: $A^{'(n)} = A^{'(n)} \times M$
    \State Scale the output feature:
      $$A^{(n+1)}=A^{'(n)} \times \texttt{count}(M) /
      \texttt{count\_ones}(M)$$
   \end{algorithmic}
\end{algorithm}

\subsection{Implementation Details of the Proposed ADGAN}

\begin{table}[ht]
\centering
\caption{Implementation Details of the Proposed ADGAN}
\includegraphics [width=3.4in]{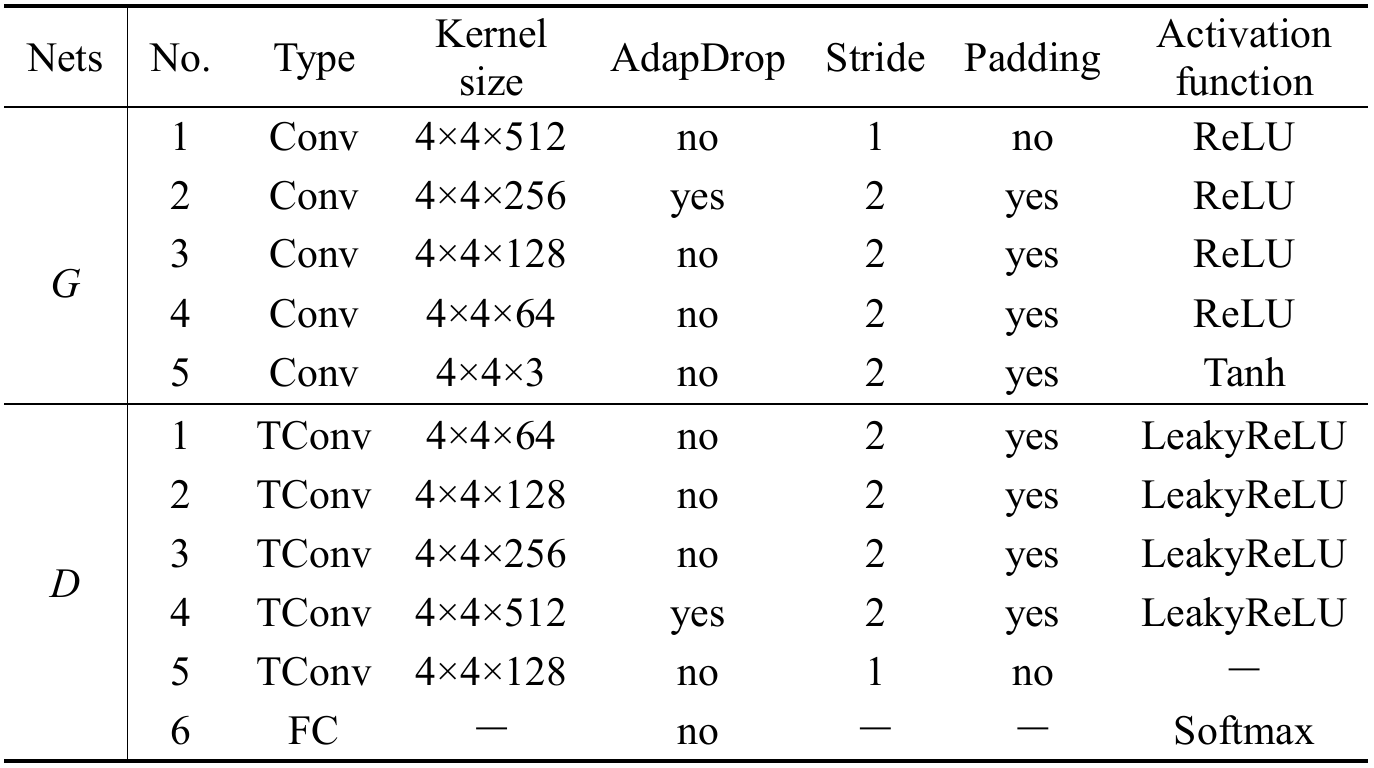}
\label{table_adgan_detail}
\end{table}

Table \ref{table_adgan_detail} shows the implementation details of the proposed ADGAN. The generator $G$ and discriminator $D$ are CNNs with five convolutional layers. The size of the input noise is $100\times1\times1$. The generator $G$ converts the inputs to fake samples with the size of $64\times64\times3$. In the generator $G$, the AdapDrop is employed in the second convolutional layer, while the AdapDrop is employed in the forth transposed convolutional layer in the discriminator $D$.

\section{Experimental Results and Analysis}

In this section, we first describe the datasets used in our experiments. Then, an exhaustive investigation of several important parameters of the AdapDrop is presented. Besides, We tested the impact of different regularization methods on classification results in the network. Next, the comparisons with five closely related HSI classifiers are provided After that, we compared the running time of different classification methods. Finally, we visualized the generated samples to show the advantage of ADGAN.

\begin{figure}[ht]
\begin{center}
\includegraphics [width=3.5in]{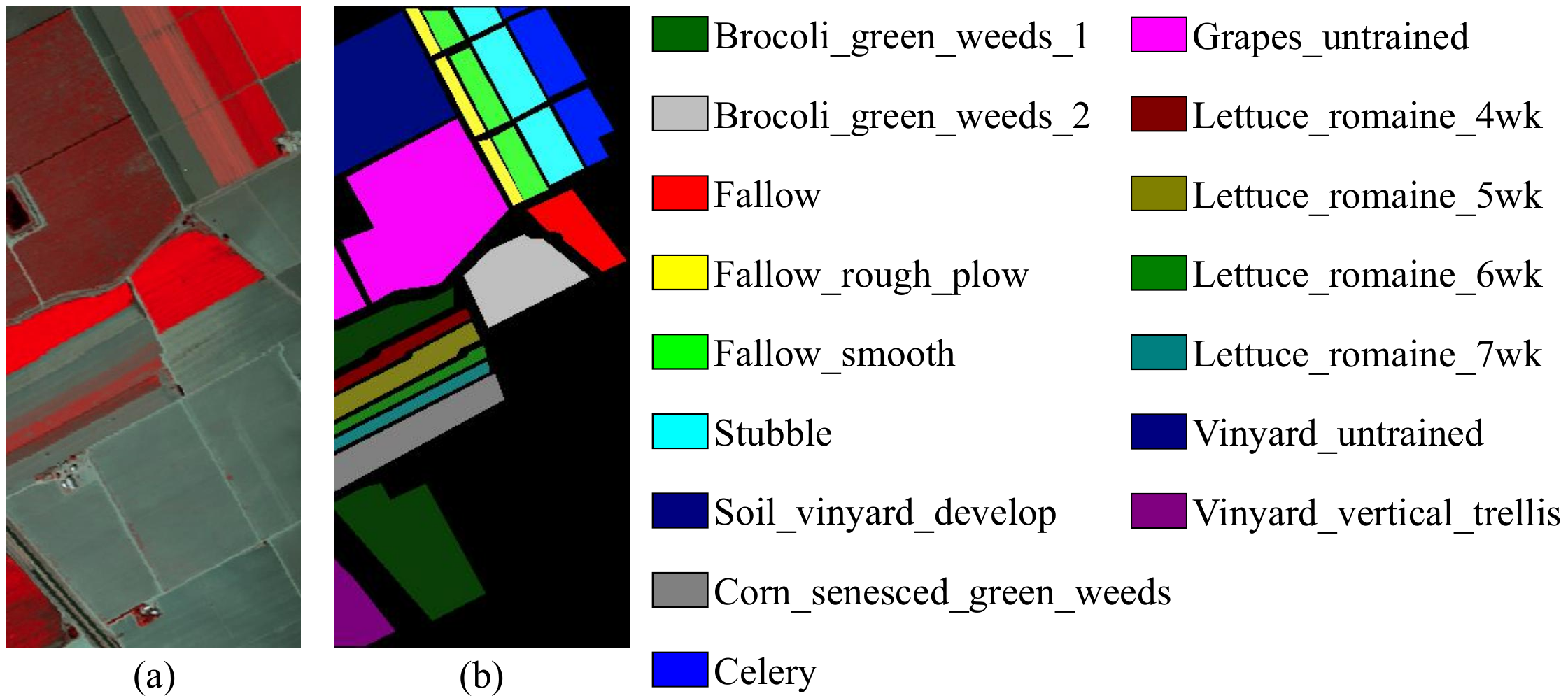}
\caption{Salinas dataset. (a) False-color composite. (b) Ground reference map.}
\label{fig_data_salinas}
\end{center}
\end{figure}

\begin{figure}[ht]
\begin{center}
\includegraphics [width=2.8in]{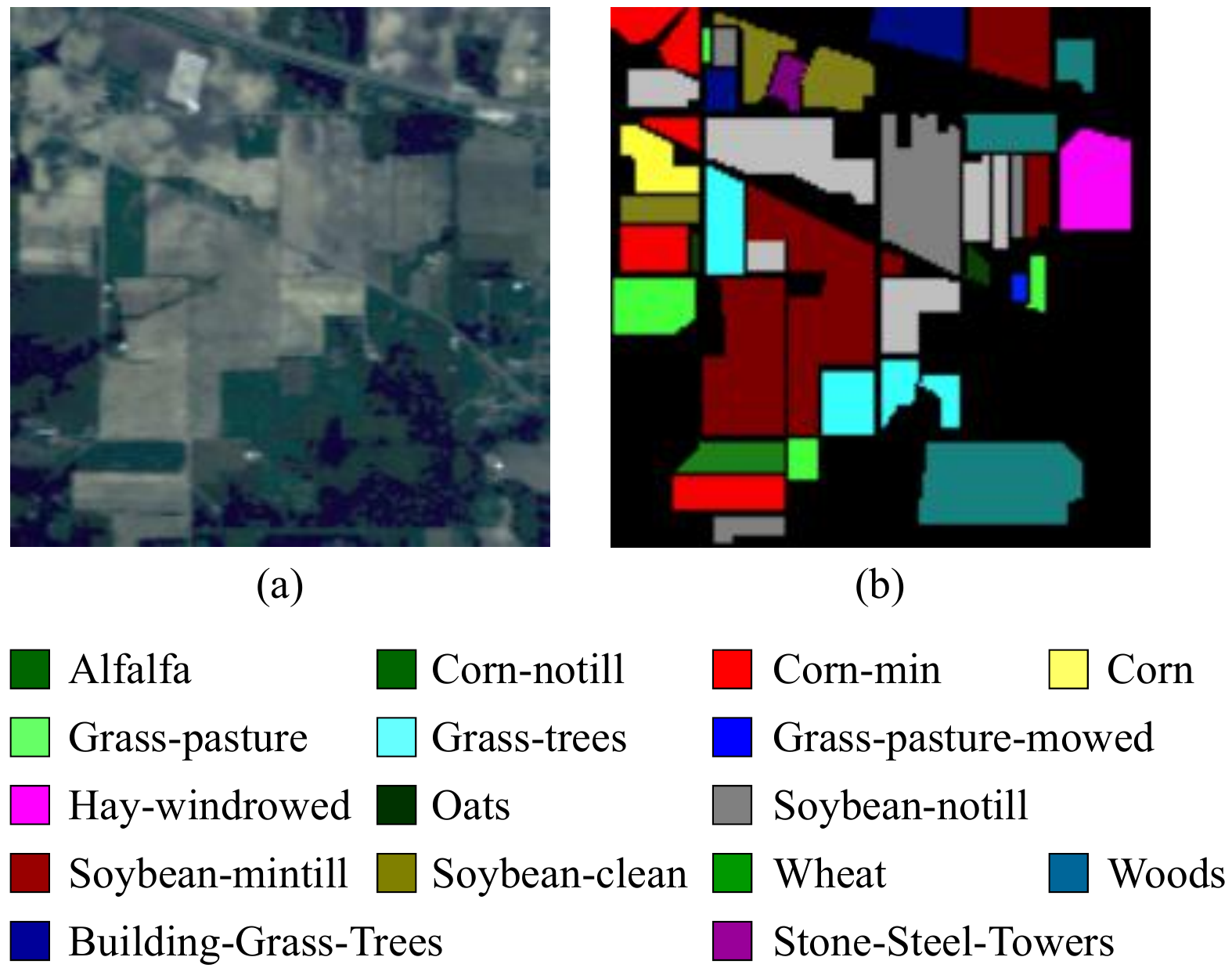}
\caption{Indian Pines dataset. (a) False-color composite. (b) Ground reference map.}
\label{fig_data_indian}
\end{center}
\end{figure}

\begin{figure}[ht]
\begin{center}
\includegraphics [width=2.8in]{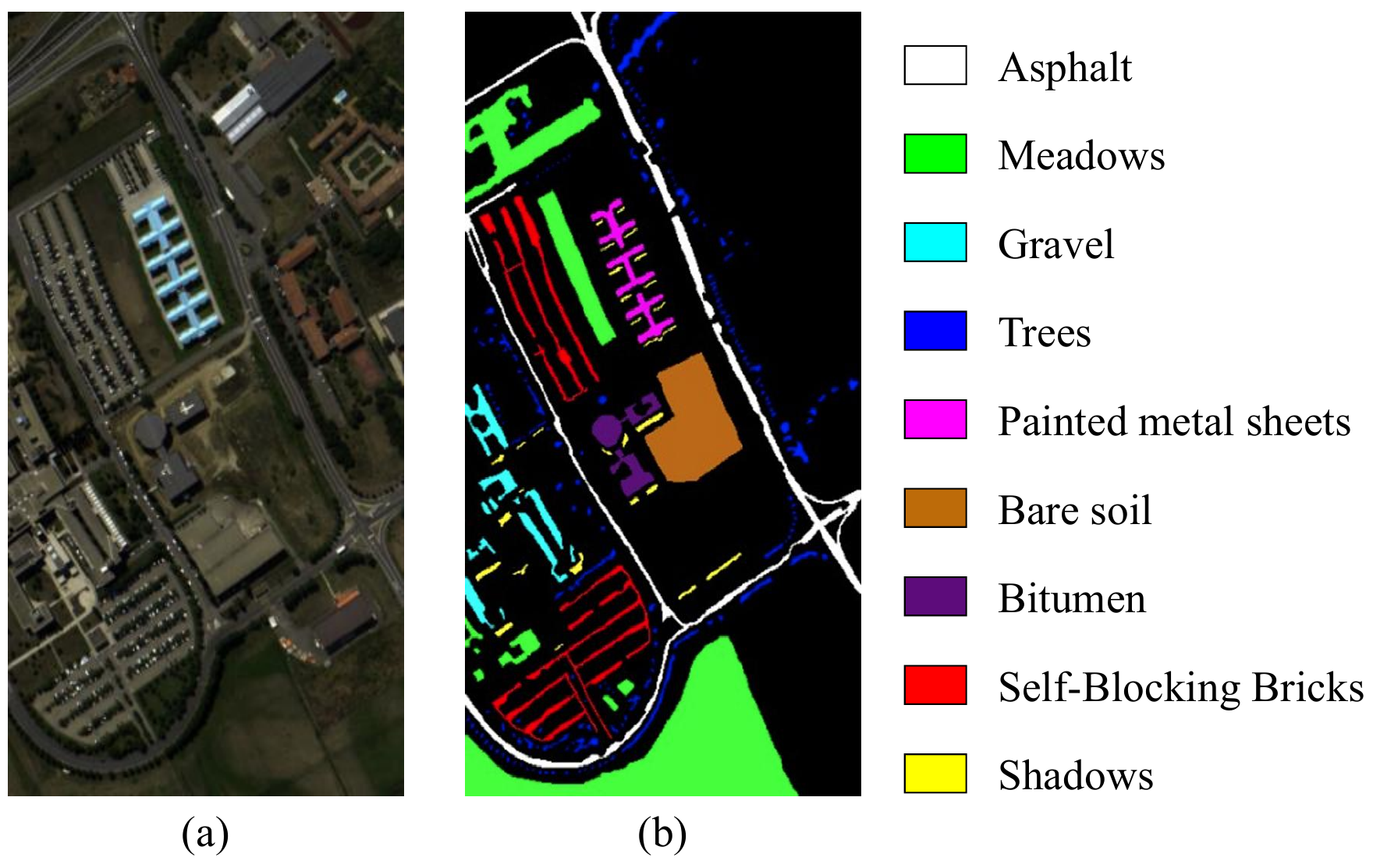}
\caption{Pavia University data set. (a) False-color composite. (b) Ground reference map.}
\label{fig_data_pavia}
\end{center}
\end{figure}

\subsection{Datasets Description}

To evaluate the performance of the ADGAN on HSI classification, three representative HSI datasets are used, including the Salinas, Indian Pines, and Pavia University datasets.

\begin{enumerate}

\item The Salinas dataset was captured by the Airborne Visible/Infrared Imaging Spectrometer (AVIRIS) sensor. The size of the image is 512$\times$217 pixels. The dataset is comprised of 204 spectral bands. Some low signal-to-noise ratio (SNR) bands are removed. The dataset has a high spatial resolution of 3.7m per pixel. The false-color composite image (bands 50, 170, 190) and the corresponding ground reference map are illustrated in Fig. \ref{fig_data_salinas}.

\item The Indian Pines dataset is a mixed vegetation site in Northwestern Indiana, and it was collected by the AVIRIS sensor. The size of the dataset is 145$\times$145 pixels. It is comprised of 220 spectral bands in the wavelength range of 0.4-2.5 $\mu$m. The false-color composite image and the ground reference map are shown in Fig. \ref{fig_data_indian}. It should be noted that the water absorption bands are removed in our implementations.

\item The Pavia University dataset was acquired by the Reflective Optics System Imaging Spectrometer (ROSIS) in northern Italy in 2001. The dataset convers nine urban land-cover types. The size of the dataset is 610$\times$340 pixels, and the resolution of the image is 1.3m per pixel. The dataset is comprised of 103 spectral bands in the wavelength range from 430 to 860 nm. Fig. \ref{fig_data_pavia} illustrates the dataset and the corresponding ground reference map.

\end{enumerate}

For all three datasets, the labeled samples were split into two parts: the training set and the test set. Because of the relatively higher computational complexity of the GANs, we try to control the number of training samples to ensure stable experimental results. After numerous experiments, we found that randomly selecting 300 training samples on the Salinas dataset, 1000 training samples on the Indian Pines dataset, and 1000 training samples on the Pavia University dataset can ensure stable results. For the Salinas dataset, the number of training and test samples for each class are listed in Table. \ref{table_sample_salinas}. For the Indian Pines and Pavia University dataset, sample distribution is listed in Table. \ref{table_sample_indian} and \ref{table_sample_pavia}, respectively. The training set adjusts the parameters during the training process by testing the classification accuracies and the losses of the temporary model generated during training. The network with the lowest loss is selected for testing. In the test process, all the test samples in the dataset are used to estimate the capability of the trained network. Three evaluation criteria, including overall accuracy (OA), average accuracy (AA), and Kappa coefficient ($\kappa$) are presented for all test samples.

\begin{table}[htbp]
\centering
\begin{center}
\caption{Samples distribution for the Salinas dataset}
\includegraphics [width=3.3in]{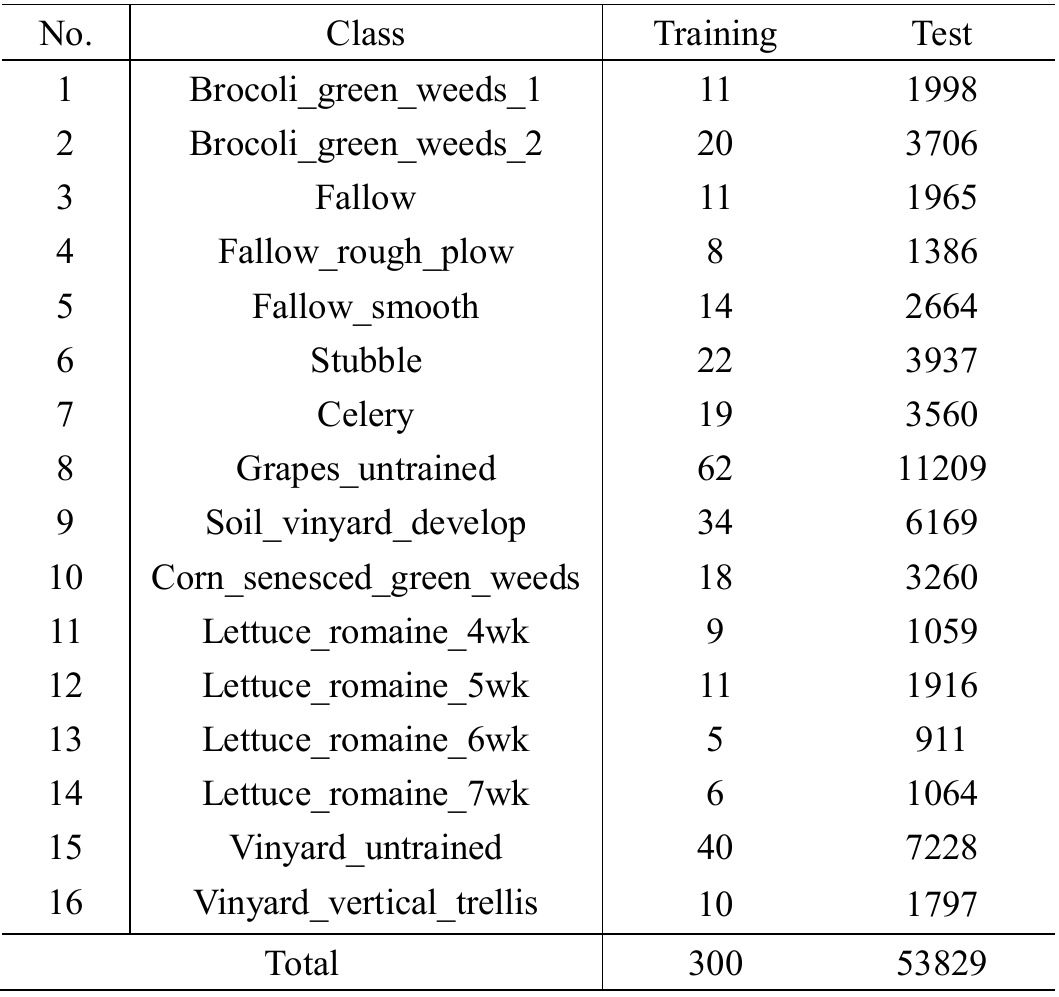}
\label{table_sample_salinas}
\end{center}
\end{table}

\begin{table}[htbp]
\centering
\begin{center}
\caption{Samples distribution for the Indian Pines dataset}
\includegraphics [width=3.3in]{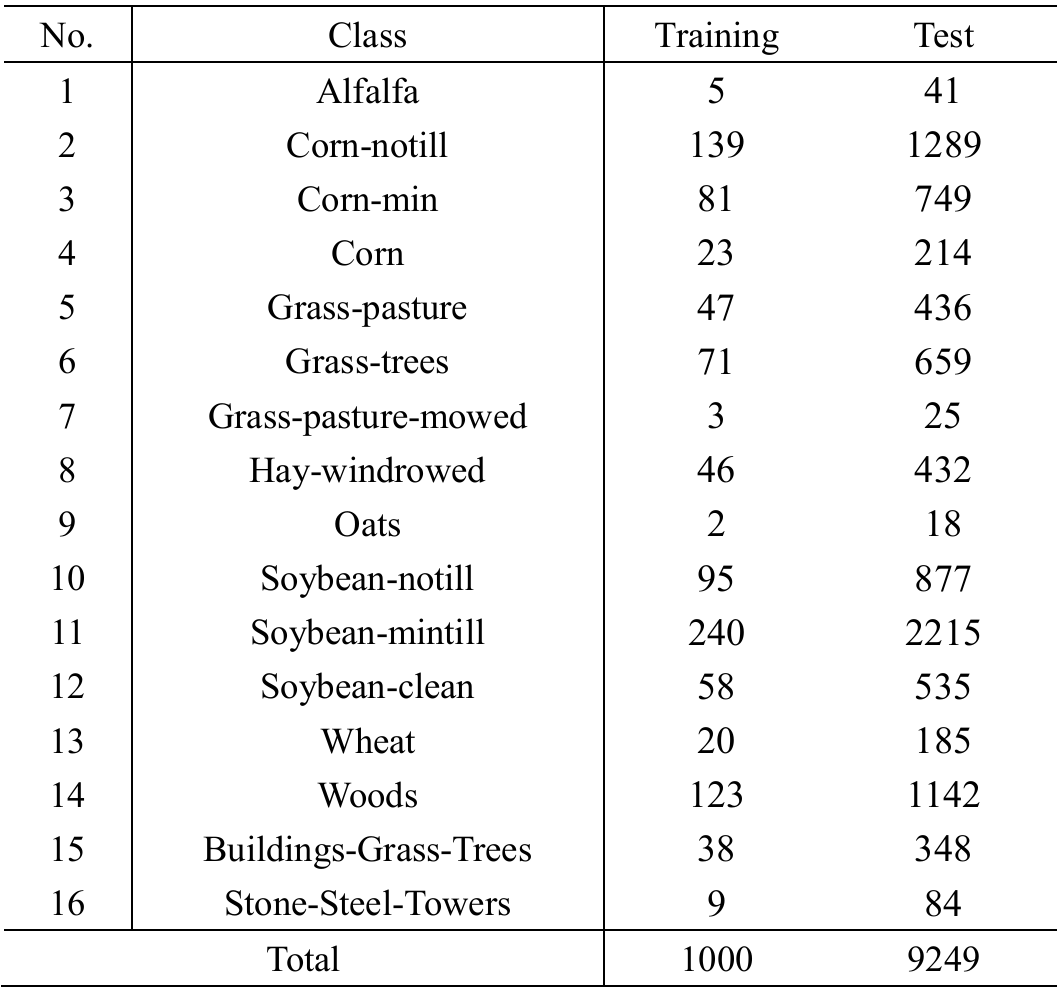}
\label{table_sample_indian}
\end{center}
\end{table}

\begin{table}[htbp]
\centering
\begin{center}
\caption{Samples distribution for the Pavia University dataset}
\includegraphics [width=3.3in]{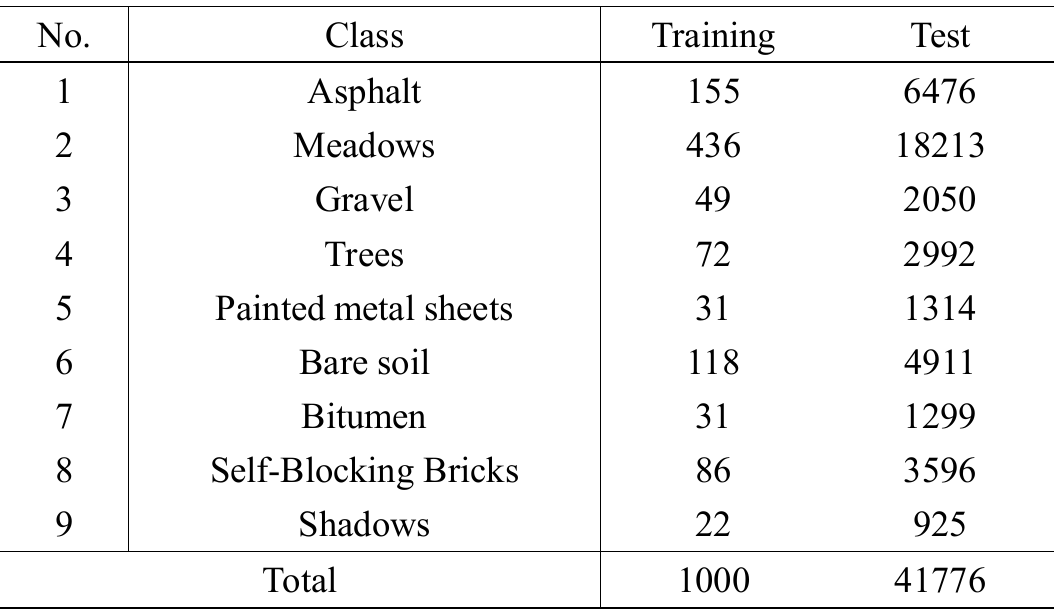}
\label{table_sample_pavia}
\end{center}
\end{table}

\subsection{Parameter Analysis}

\subsubsection{Analysis of $b\_size$}

\begin{figure}[ht]
\centering
\includegraphics [width=2.8in]{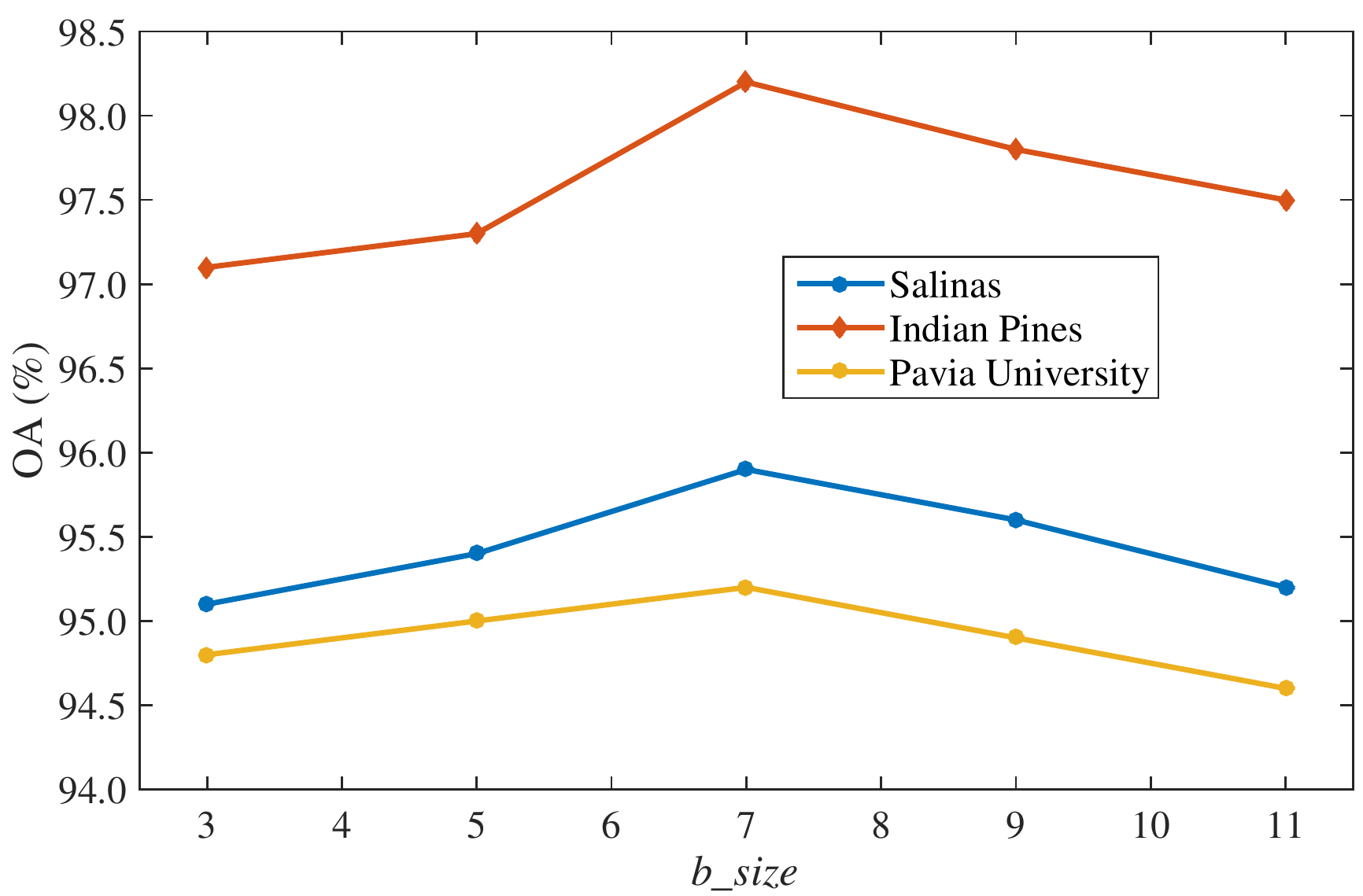}
\caption{Relationship between the classification accuracy and $b\_size$ on three datasets.}
\label{fig_bsize}
\end{figure}

The size of block $b\_size$ in AdapDrop is an important parameter that affects the classification accuracy. The contextual information in classification is sensitive to neighborhood noise. Fig. \ref{fig_bsize} shows the classification performance on three datasets under different $b\_size$. In our implementations, the $b\_size$ varies from 3 to 11. We can observe that when $b\_size$ increases from 3 to 7, the classification accuracy improves since more contextual information is taken into account. However, when the larger block size is selected, continuous blank areas affect the robust training of the network. When $b\_size=7$, the best accuracy is achieved. Therefore, $b\_size$ is set to 7 in the following experiments.

\subsubsection{Analysis of $k$}

\begin{figure}[ht]
\centering
\includegraphics [width=2.8in]{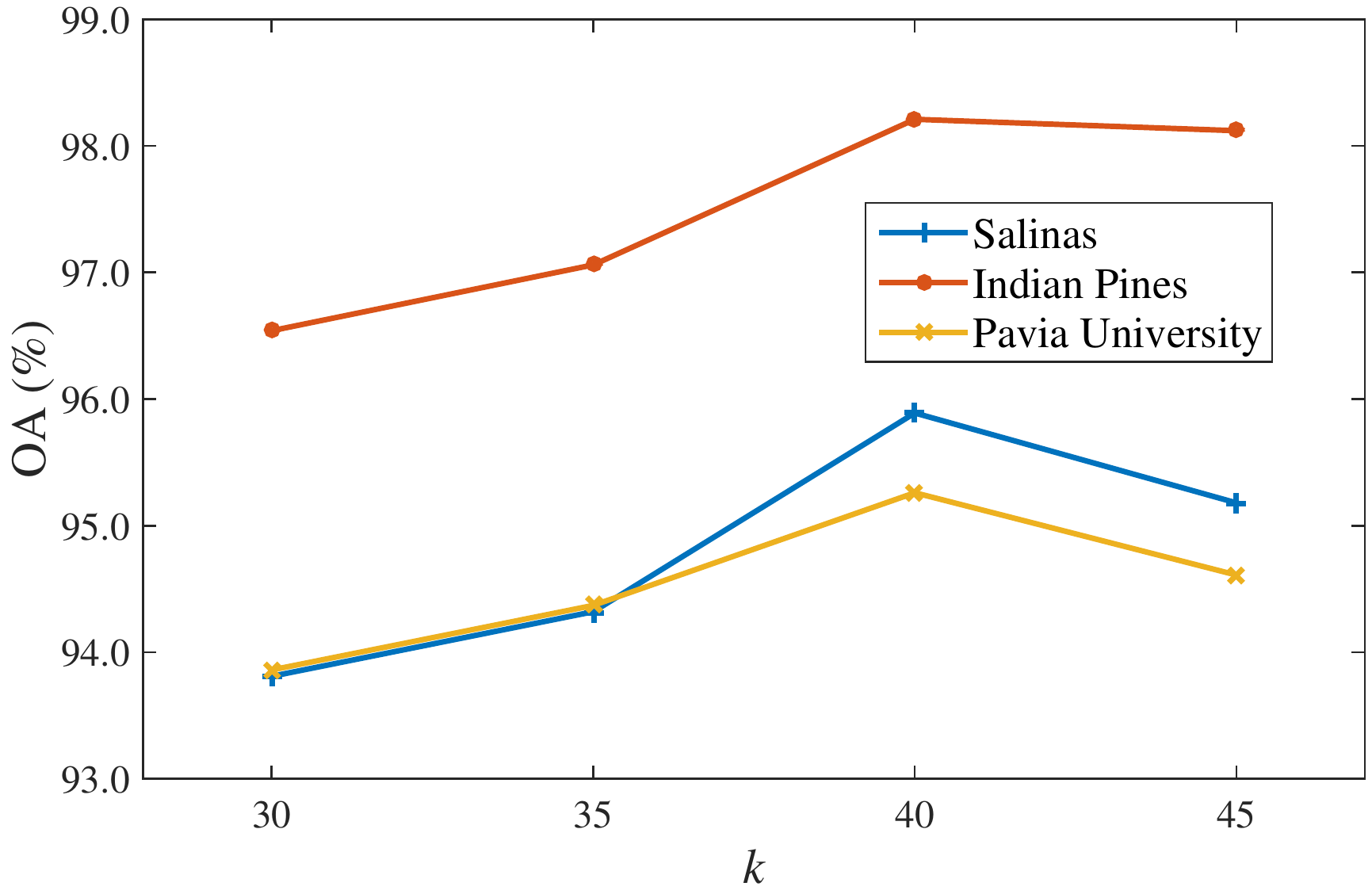}
\caption{Relationship between the classification accuracy and $k$ on three datasets.}
\label{fig_k}
\end{figure}

The parameter $k$ denotes that the top $k$th percentile elements in the mask block will be dropped, and it is a critical parameter in AdapDrop. We evaluate the classification performance by take $k$ = 30, 35, 40, and 45, respectively. Fig. \ref{fig_k} illustrates the influence of $k$ on classification accuracy on three datasets. It can be seen that when $k=40$, the OA achieves the best performance. When smaller $k$ is selected, the dropped features can hardly achieve the goal of mitigating overfitting. When it gets bigger, the network dropped too much features, and the network inclines to learn incorrect representations provided by the irrelevant background. Therefore, $k$ is chosen as 40 in our following experiments.

\begin{figure}[ht]
\centering
\includegraphics [width=2.8in]{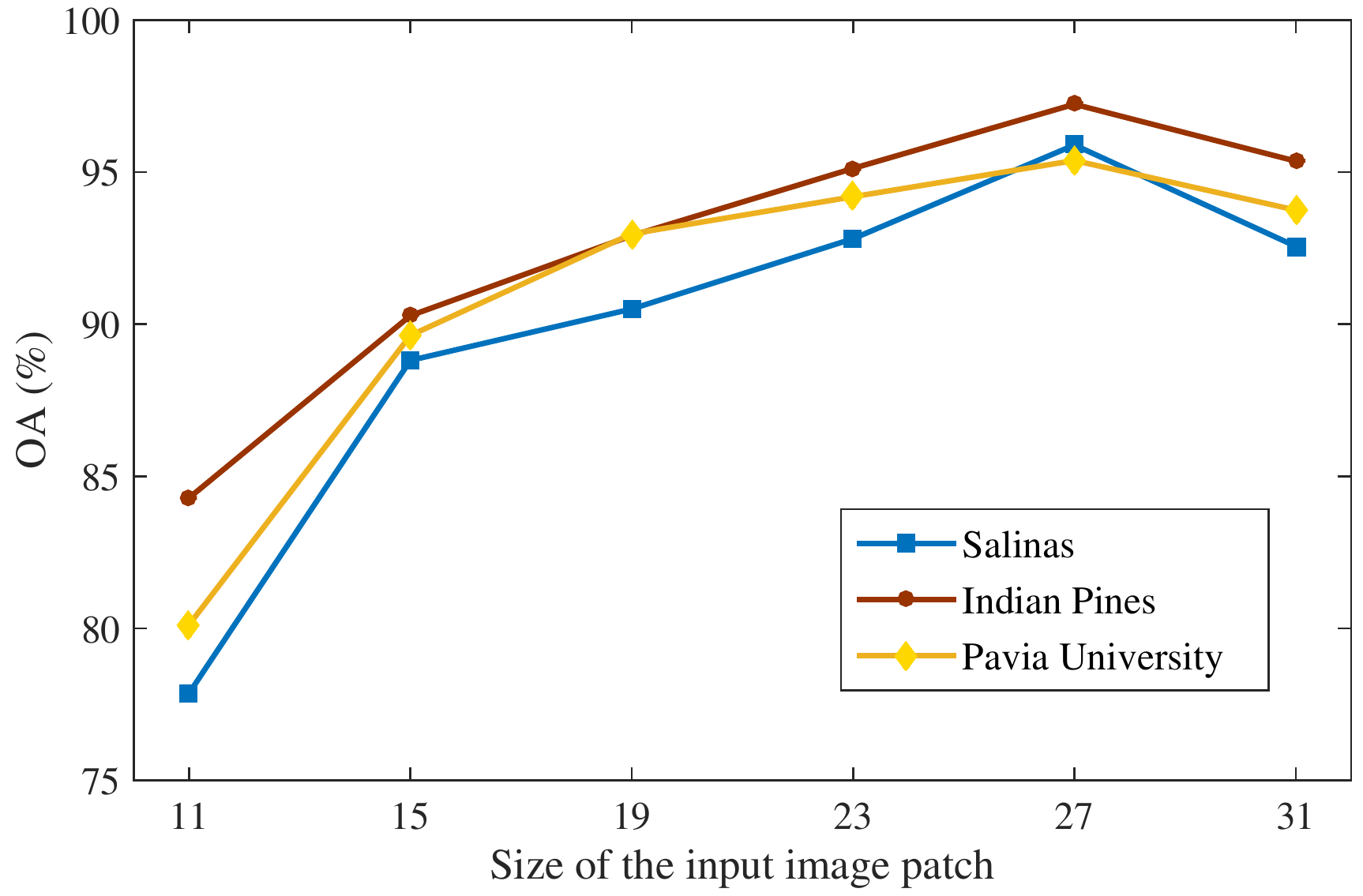}
\caption{Classification performance versus different input image patch sizes on three datasets.}
\label{fig_patch_size}
\end{figure}

\begin{figure}[ht]
\centering
\includegraphics [width=2.8in]{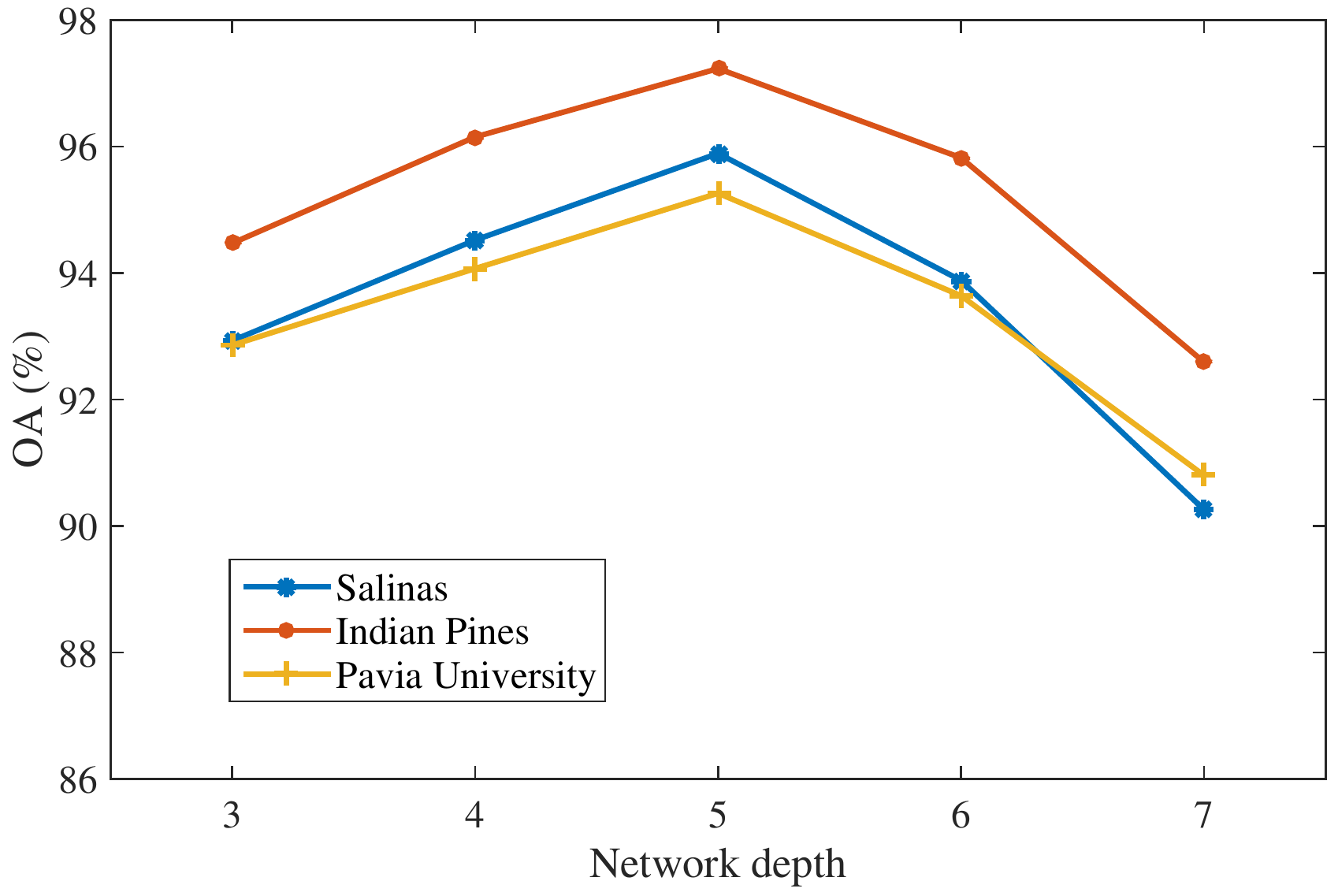}
\caption{Classification performance versus network depths on three datasets.}
\label{fig_network_depth}
\end{figure}

\subsubsection{Analysis of the size of the patch}

The size of the input image patch is an important parameter. As illustrated in \ref{fig_patch_size}, the input image patches are set to 11, 15, 19, 23, 27, and 31, respectively. It can be observed that the classification accuracy sharply increases when the patch size ranges from 11 to 27 on three datasets. When patch size grows larger than 27, the classification accuracy tends to decrease. It is owing to the reason that larger image patch takes pixels of different classes into account, and hence some negative effects are incurred. In the meanwhile, the valuable spatial information is not exploited effectively when the patch size is rather small. Therefore, the extracted features are not representative of the central pixel. Therefore, in our implementations, the input image patch size is set to 27 $\times$ 27 pixels on three datasets.

\subsubsection{Analysis of network depth}

It is widely acknowledged that the network depth of current deep learning-based methods is getting deeper and deeper. However, when training samples are relatively limited, the parameters in deeper models can hardly be optimized, and the model is unable to work well. As illustrated in Fig. \ref{fig_network_depth}, when the network depth is set to 5, the best performance is obtained on three datasets. It is reasonable that deeper architectures may suffer from the exploding gradients problem. Specifically, error gradients accumulate quickly and thus result in an unstable network. Hence, the network depth is set to be 5.

\subsection{Comparison with Different Regularization Methods}

\begin{figure}[ht]
\centering
\includegraphics [width=2.8in]{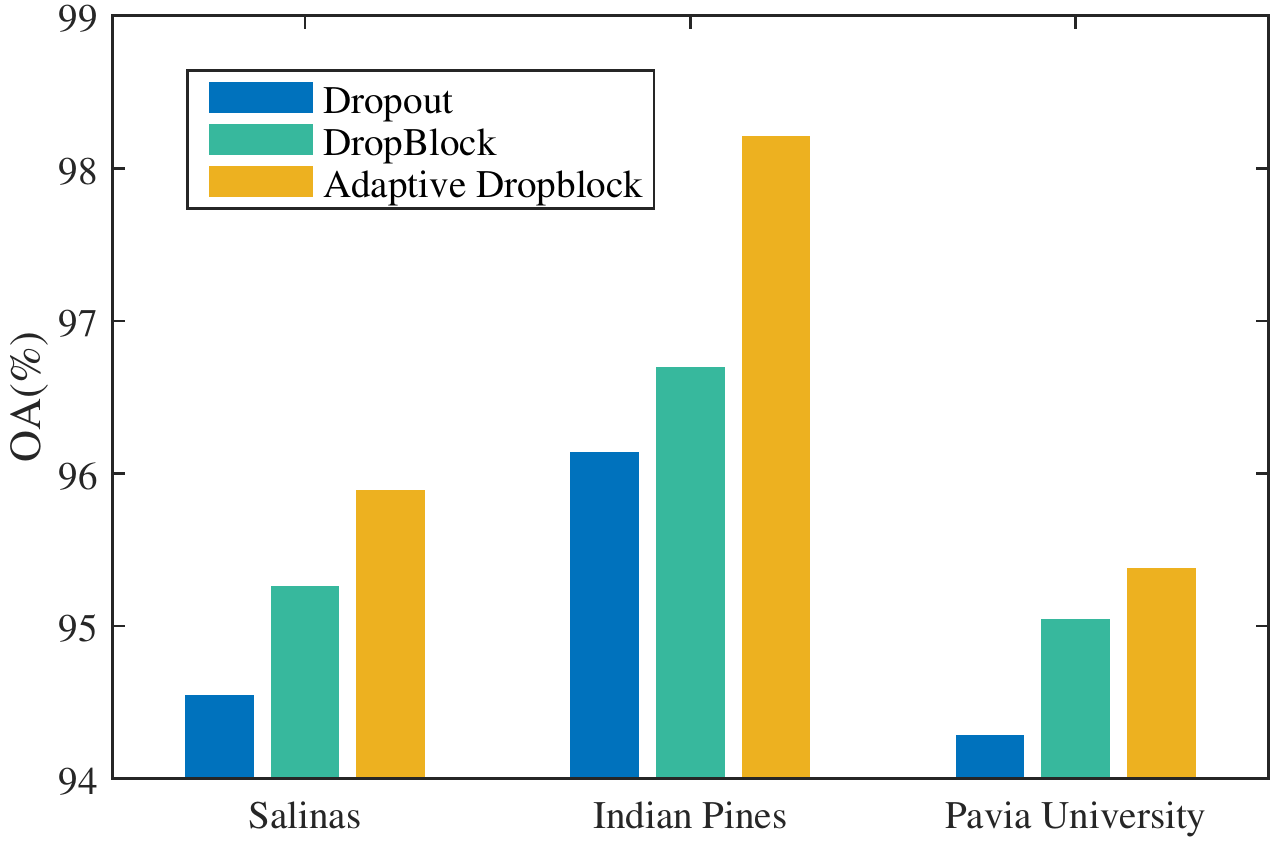}
\caption{Classification accuracy by employing different regularization methods on three datasets.}
\label{fig_drop}
\end{figure}

In this subsection, we empirically investigate the effectiveness of the proposed AdapDrop for HSI classification. Dropout and DropBlock are employed for comparison in extensive experiments. As shown in Fig. \ref{fig_drop}, the proposed AdapDrop has a superior performance compared with Dropout and DropBlock. Among these methods, Dropout is less effective since it randomly drop separate pixels in the feature map, and the dropped information can be easily retrieved through neighborhood pixels. DropBlock removes the entire blocks, and the network's learning capabilities may be affected. The proposed AdapDrop removes highly informative regions in the feature map, and the network can effectively learn robust featuresof the ground objects in HSI classification.

\begin{figure}[htb]
\centering
\includegraphics [width=3.0in]{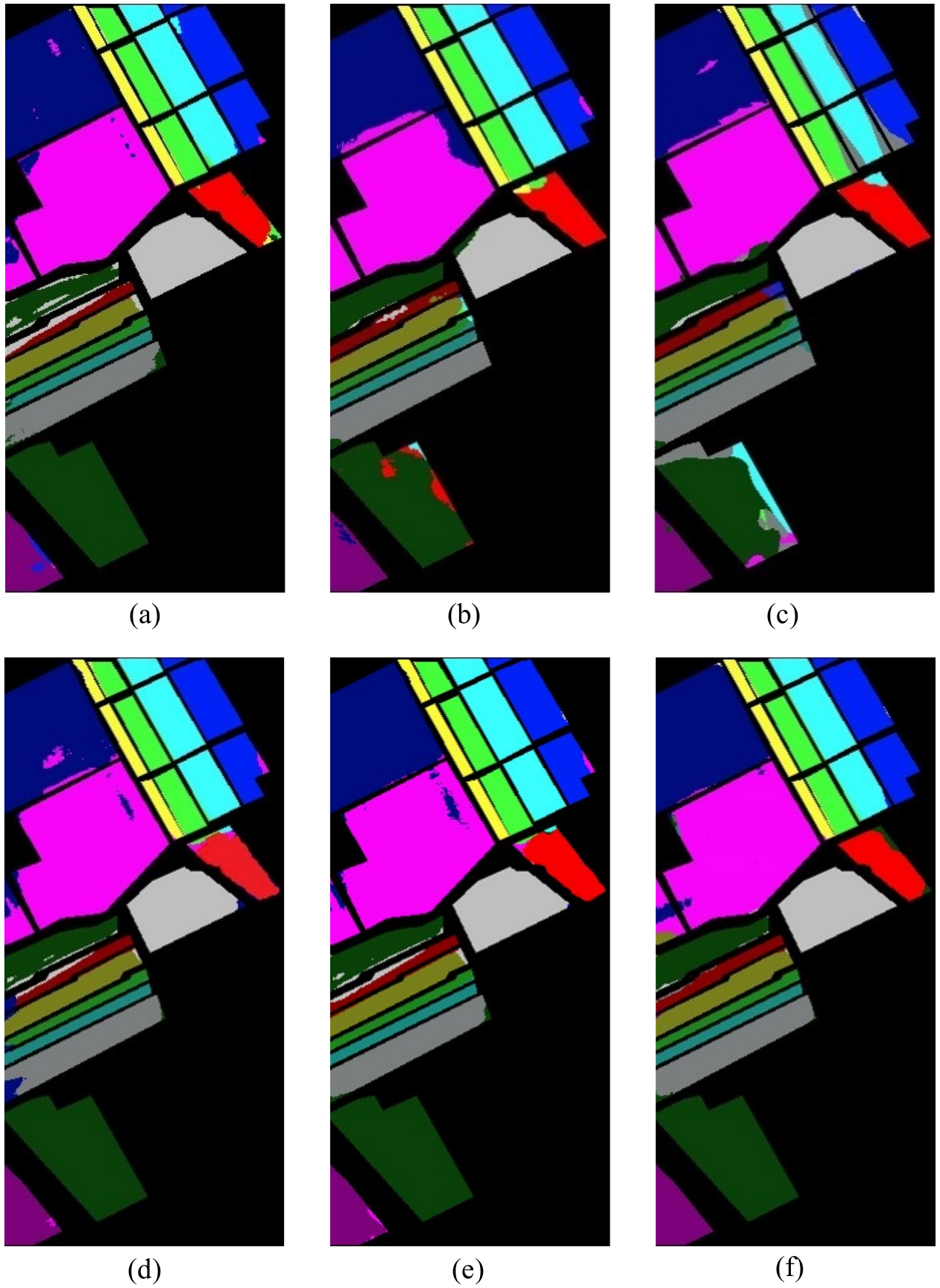}
\caption{Visualized results of different classification methods on the Salinas dataset. (a) Result by RF \cite{Ham05_tgrs}. (b) Result by CSVM \cite{Gurram13_grsl}. (c) Result by EP-CNN \cite{Ghamisi17_jstars}. (d) Result by SS-ResNet \cite{Haut19_tgrs}. (e) Result by 3D-ACGAN \cite{Zhu18_tgrs}. (f) Result by the proposed ADGAN.}
\label{fig_res_salinas}
\end{figure}

\begin{figure}[htb]
\centering
\includegraphics [width=3.5in]{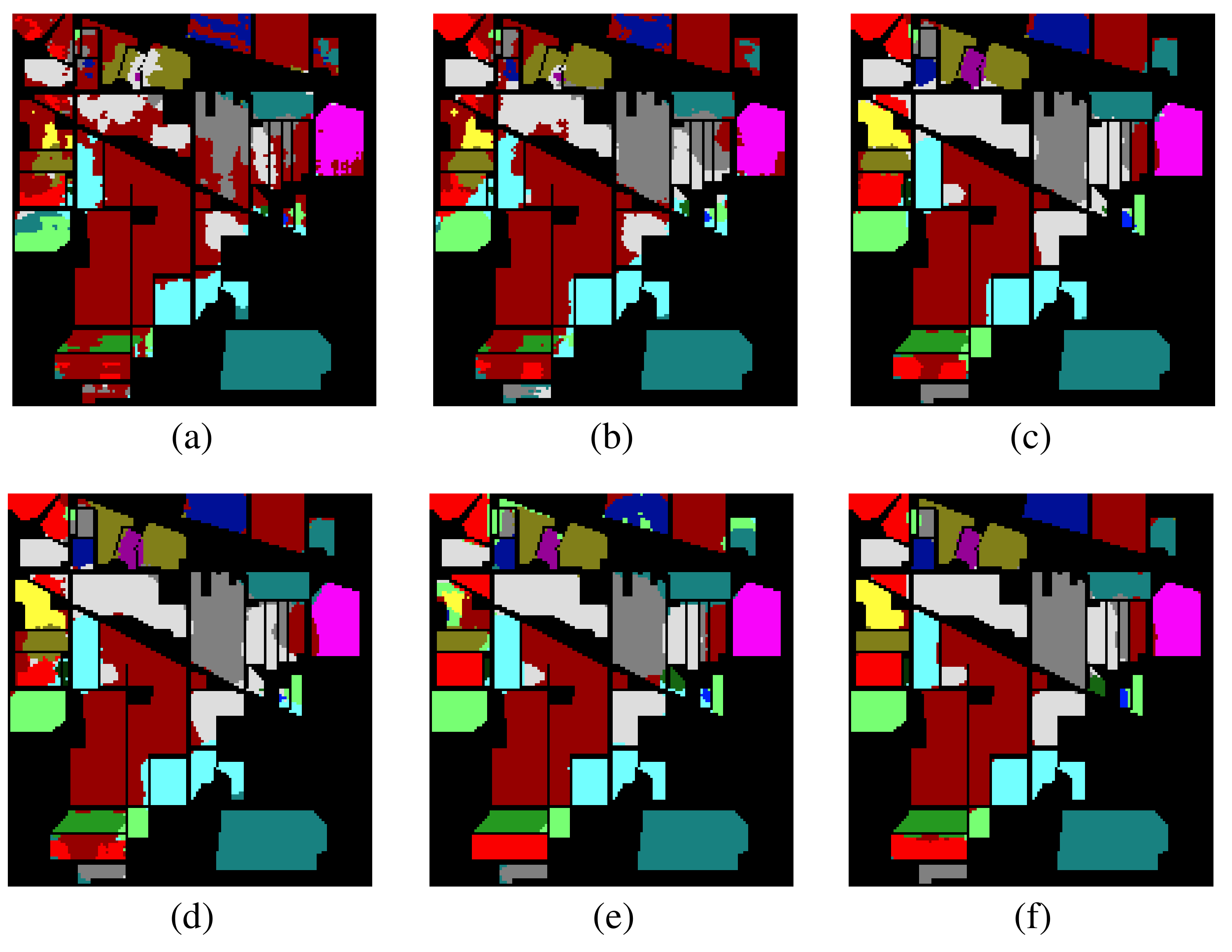}
\caption{Visualized results of different classification methods on the Indian Pines dataset. (a) Result by RF \cite{Ham05_tgrs}. (b) Result by CSVM \cite{Gurram13_grsl}. (c) Result by EP-CNN \cite{Ghamisi17_jstars}. (d) Result by SS-ResNet \cite{Haut19_tgrs}. (e) Result by 3D-ACGAN \cite{Zhu18_tgrs}. (f) Result by the proposed ADGAN.}
\label{fig_res_indian}
\end{figure}

\subsection{Classification Results on Three Datasets}

In order to verify the effectiveness of the proposed ADGAN, we compare it with five closely related methods such as the random forest (RF) \cite{Ham05_tgrs}, contextual SVM (CSVM) \cite{Gurram13_grsl}, CNN with extinction profiles (EP-CNN) \cite{Ghamisi17_jstars}, spectral-spatial ResNet (SS-ResNet) \cite{Haut19_tgrs}, and 3D-ACGAN \cite{Zhu18_tgrs}. In order to ensure a fair comparison, all the methods use default parameters, and the same proportion of training sets. All the experimental results are obtained by running 10 times independently with a random division for training and test sets.

RF investigate a random forest of binary classifiers as a means of increasing diversity of hierarchical classifiers. The $N_f$ is set to be 20, and one hundred trees are grown for each experiment. For CSVM, both local spectral and spatial information in a reproducing kernel Hillbert space are jointly exploited. A neighborhood of 9$\times$9 pixels is employed, and default parameters of SVM are used as mentioned in \cite{Gurram13_grsl}. EP-CNN fuses the hyperspectral and light detection and ranging-derived data using extinction profiles and deep learning. A neighborhood with the size of 27$\times$27 pixels is considered. $\alpha=3$ and $s=7$ are employed as described in \cite{Ghamisi17_jstars}. The SS-ResNet combines spatial and spectral information, and it takes advantage of residual learning. A neighbor hood with the size of 11$\times$11 is employed. In addition, 300 epochs and the Adam optimizer are used. For 3D-ACGAN, the source code provided by Prof. Chen is used and default parameters are chosen as mentioned in \cite{Zhu18_tgrs}. Specifically, 64$\times$64 neighborhood of each pixel is used, and the input images are normalized into the range [-0.5, 0.5]. The size of mini-batch is 100, and the Adam optimizer is employed. For data preprocessing, 3 components are utilized as the inputs. The generator $G$ and discriminator $D$ are designed with 5 convolutional layers. The size of the input noise is 100$\times$1$\times$1, and the generator converts the inputs to fake samples with the size of 64$\times$64$\times$3. In order to fairly compare the proposed ADGAN with 3D-ACGAN, both methods have similar architectures except for the output of the discriminator.

Both visual and quantitative analyses are provided in our experiments. For visual analysis, the classification maps generated by different methods are illustrated in figure form. For quantitative analysis, the classification maps are illustrated in tabular form.

\subsubsection{Results on the Salinas Dataset}

Table \ref{table_res_salinas} lists the corresponding evaluation criteria of six algorithms. The first 16 rows illustrate the results of each class, and the last 3 rows show the OA, AA and Kappa coefficients. The best classification results are emphasized by bolding. As shown in Table \ref{table_res_salinas}, deep learning-based methods, EP-CNN, SS-ResNet, 3D-ACGAN and ADGAN, are superior to RF and CSVM because of the hierarchical nonlinear feature extraction. Compared with EP-CNN and SS-ResNet, 3D-ACGAN improves the classification performance with the assistance of generated samples. Among the six methods, ADGAN achieves the best classification results in most cases since it not only generated high-quality samples but also alleviated the drawback of ACGAN as mentioned before. Additionally, compared with other methods, ADGAN improves at least 1.11\% in OA, 0.79\% in AA and 0.41\% in Kappa. Fig. \ref{fig_res_salinas} shows the classification maps of different methods on the Salinas dataset. As illustrated in Fig. \ref{fig_res_salinas}(a)-(d), RF, CSVM, EP-CNN and SS-ResNet misclassify many samples at the boundary of different classes. Compare with these methods, 3D-ACGAN achieves better classification result on majority classes because of data augmentation. Compared with 3D-ACGAN, the proposed ADGAN performs better in minority classes, for example Lettuce\_romaine\_4wk, Lettuce\_romaine\_6wk, Lettuce\_romaine\_7wk. It is demonstrated that the proposed ADGAN achieves the best performance on the Salinas dataset.

\begin{table*}[htbp]
\centering
\begin{center}
\caption{Classification results obtained by different classification methods on the Salinas data set}
\includegraphics [width=6.0in]{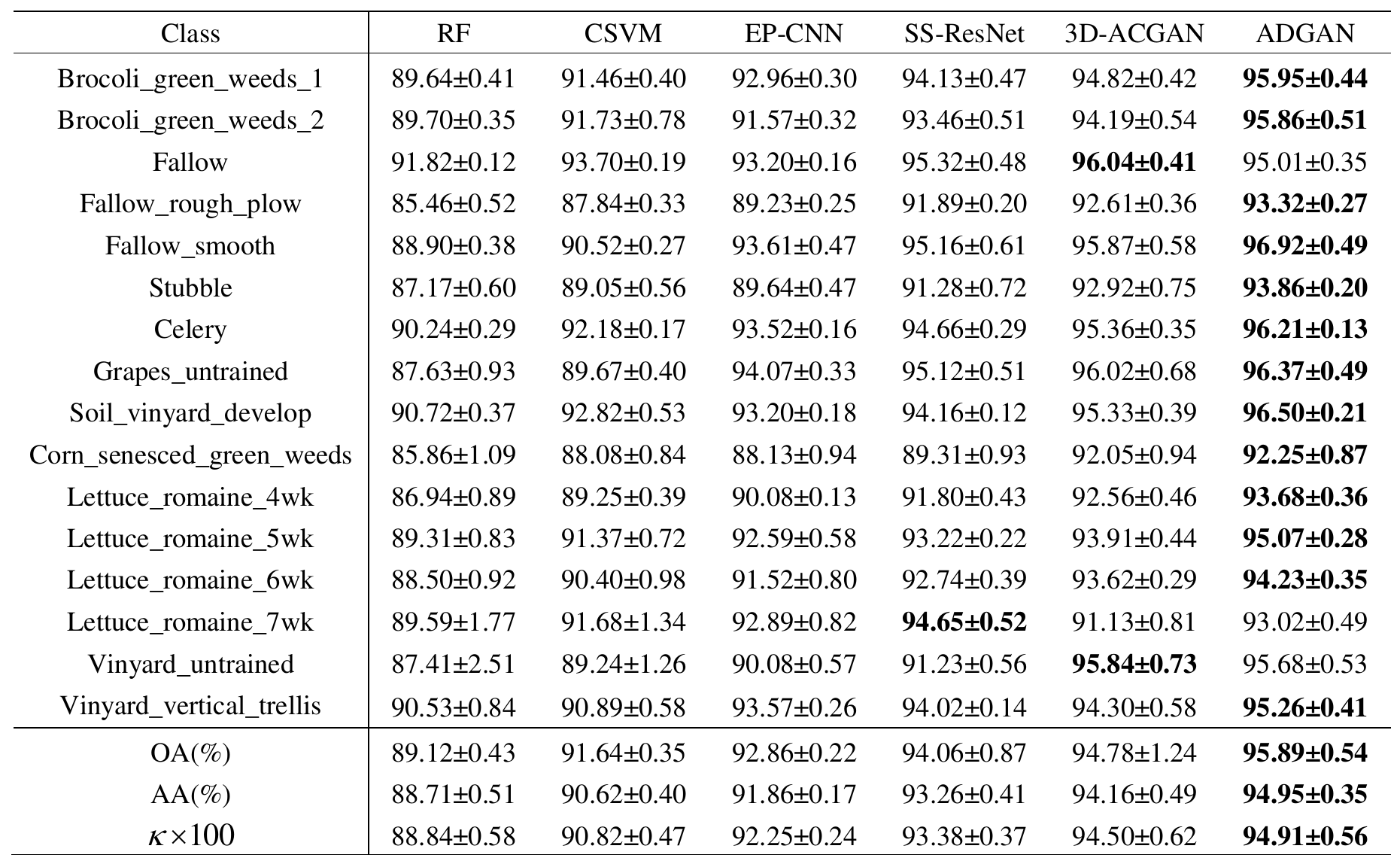}
\label{table_res_salinas}
\end{center}
\end{table*}

\subsubsection{Results on the Indian Pines dataset}
The statistical classification results on the Indian Pines dataset are summarized in Table \ref{table_res_indian}, and Fig. \ref{fig_res_indian} illustrates the classification results of different methods. As can be observed from Table \ref{table_res_indian}, SS-ResNet, 3D-ACGAN and ADGAN are superior to RF, CSVM and EP-CNN by introducing attention mechanism or extra generated training samples. For minority classes, such as Alfalfa, Grass-pasture-mowed, Oats and Stone-Steel-Towers, the classification performance of the proposed ADGAN is better than 3D-ACGAN. It is demonstrated that ADGAN has better classification performance when handling minority class samples on this dataset. Among all these methods, ADGAN obtains the best statistical results in terms of the OA, AA and Kappa. As shown in Fig. \ref{fig_res_indian}(a)-(d), many samples belonging to the Soybean-clean and Building-Grass-Trees are falsely assigned the neighboring labels by RF, CSVM, EP-CNN and SS-ResNet. Compared with them, ADGAN achieves better region uniformity in the Soybean-notill class. Moreover, ADGAN obtains better performance in boundary pixel classification of the Stone-Steel-Towers, which is limited in training set. It is evident that the proposed ADGAN obtains the best performance on the Indian Pines dataset.

\begin{table*}[htbp]
\centering
\begin{center}
\caption{Classification results obtained by different classification methods on the Indian Pines data set}
\includegraphics [width=6.0in]{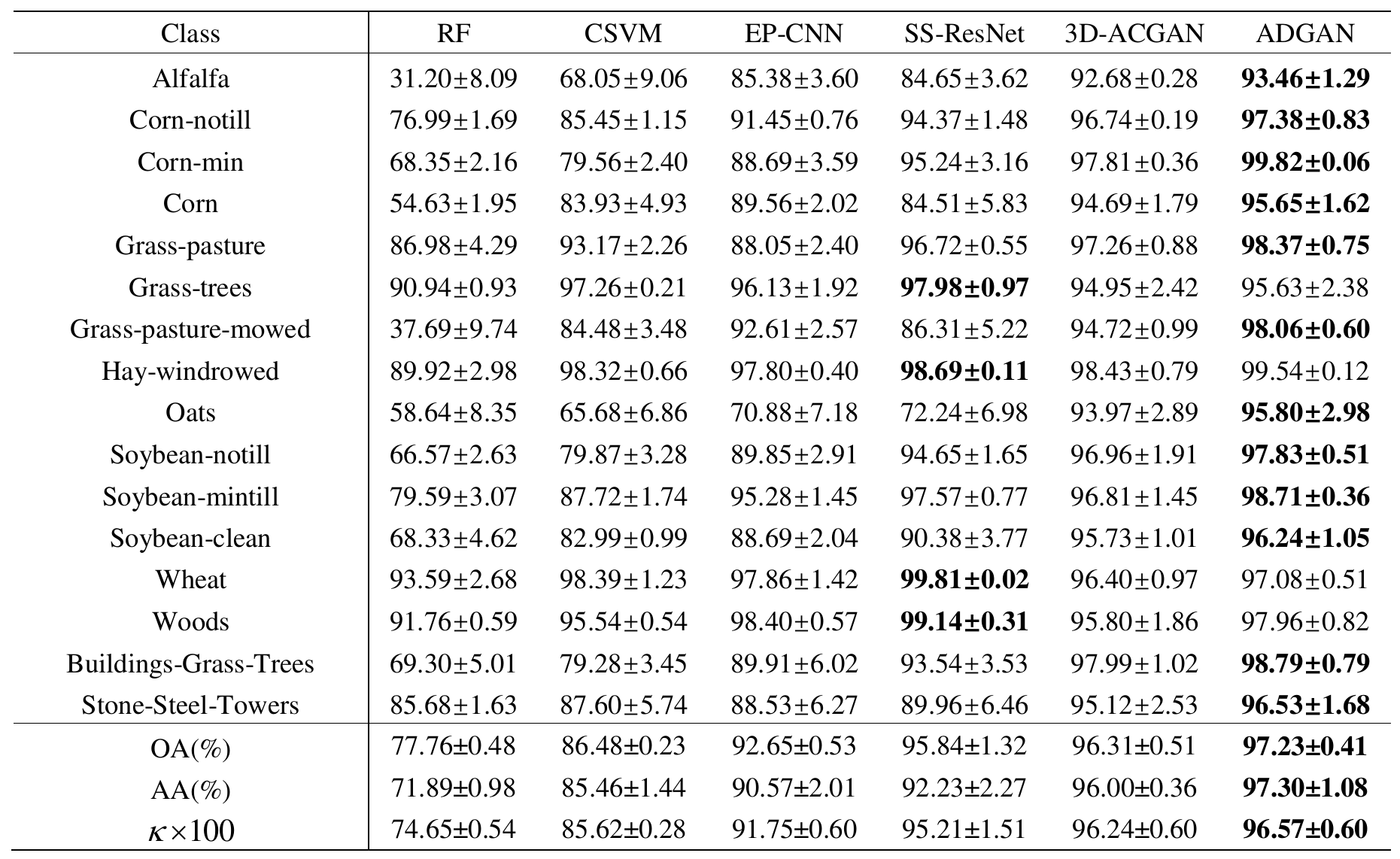}
\label{table_res_indian}
\end{center}
\end{table*}

\subsubsection{Results on the Pavia University dataset}

\begin{figure}[ht]
\centering
\includegraphics [width=3.5in]{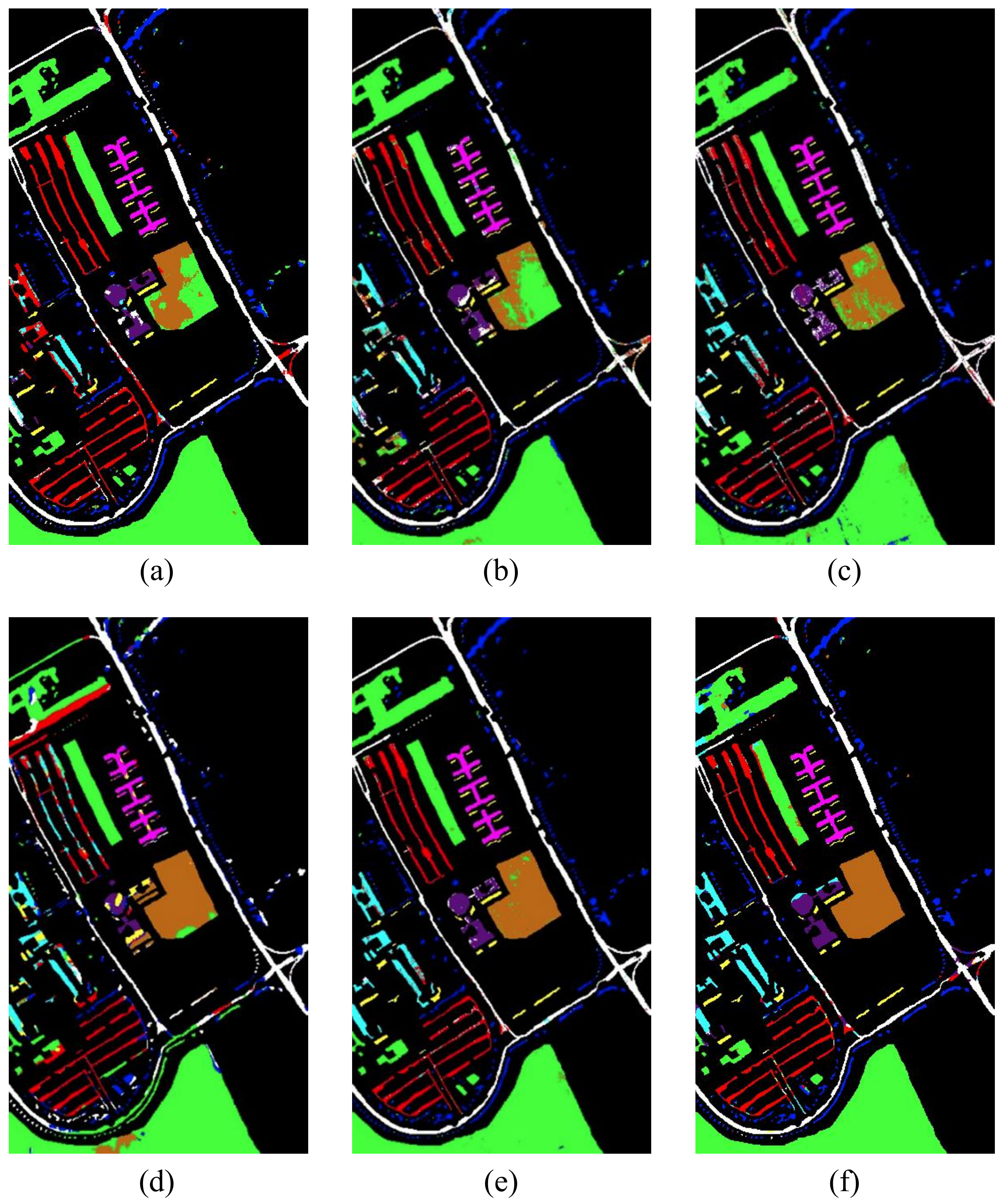}
\caption{Visualized results of different classification methods on the Pavia University dataset. (a) Result by RF \cite{Ham05_tgrs}. (b) Result by CSVM \cite{Gurram13_grsl}. (c) Result by EP-CNN \cite{Ghamisi17_jstars}. (d) Result by SS-ResNet \cite{Haut19_tgrs}. (e) Result by 3D-ACGAN \cite{Zhu18_tgrs}. (f) Result by the proposed ADGAN.}
\label{fig_res_pavia}
\end{figure}

\begin{table*}[htbp]
\centering
\begin{center}
\caption{Classification results obtained by different classification methods on the Pavia University data set}
\includegraphics [width=6.0in]{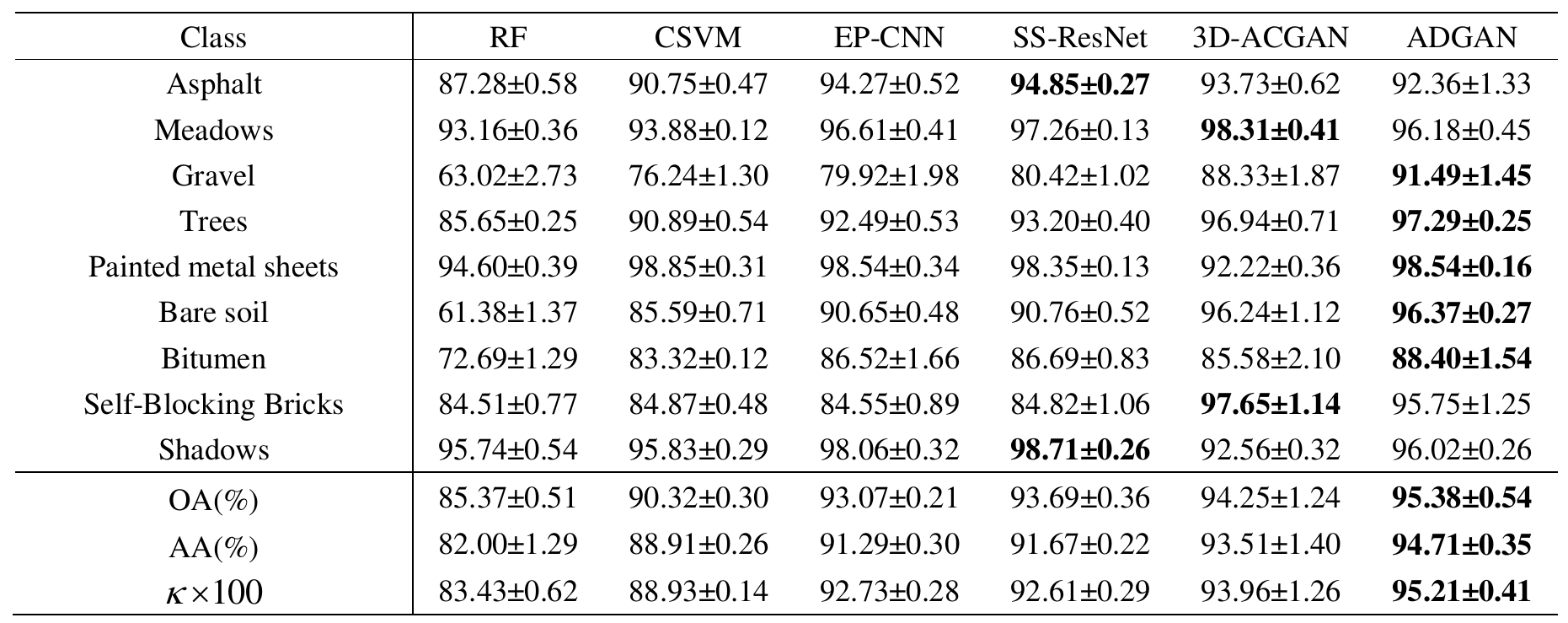}
\label{table_res_pavia}
\end{center}
\end{table*}

The quantitative criteria of different methods on the Pavia University dataset are shown in Table \ref{table_res_pavia}. The corresponding classification maps on the dataset are illustrated in Fig. \ref{fig_res_pavia}. As can be observed in Table \ref{table_res_pavia}, the Gravel and Self-Blocking Bricks classes are misclassified by RF, CSVM, EP-CNN and SS-ResNet. Compared with these methods, 3D-ACGAN and ADGAN obviously improve the classification performance by generating high-quality training samples. When handling the minority classes, such as Painted metal sheets, Bitumen and Shadows, the proposed ADGAN performs better than 3D-ACGAN. From visual comparisons, the proposed ADGAN obtains the best classification results. The proposed ADGAN surpasses 3D-ACGAN by 1.13\%, 1.20\% and 1.25\% in terms of OA, AA and Kappa. As shown in Fig. \ref{fig_res_pavia}(a)-(d), there are many noisy scattered points in the Bare soil and Gravel in the classification results by RF, CSVM, EP-CNN and SS-ResNet. Compared to them, 3D-ACGAN and ADGAN provide better results with little noise. It should be noted that, when handling the minority classes, such as Bitumen and Shadows, ADGAN performs better and is the closest to the ground truth map. The experimental results on this dataset demonstrate that the ADGAN exhibits good classification performance by capturing the intrinsic inter-class discriminative features.

From visual comparisons, the classification results by the proposed ADGAN are less noisy than the other methods. The quantitative criteria in Tables \ref{table_res_salinas} - \ref{table_res_pavia} are consistent with the visual comparisons. It should be noted that deep learning-based methods generally perform better than shallow architectures. Especially, the GAN-based methods indeed obtain better classification results when the training samples are limited. The proposed ADGAN is capable to achieve better classification accuracy than 3D-ACGAN in minority class classification owing to the newly designed discriminator and AdapDrop.

\subsection{Investigation on Running Time}

Table \ref{time} lists the running time of different classification methods on three datasets. Compared with RF and CSVM, deep learning-based methods cost more training time because of the construction of deep network. 3D-ACGAN and ADGAN are time-consuming on the training time because adversarial learning needs more time to converge. For the test time, the proposed ADGAN has obvious advantage than EP-CNN and SS-ResNet because of the simpler network structure of the discriminator. Furthermore, we can observe that ADGAN has competitive performance compared with 3D-ACGAN in test time. It means that the proposed ADGAN is capable of real-time applications.

\begin{table}[ht]
\caption{Running time of different classification methods on three data sets}
\includegraphics [width=3.4in]{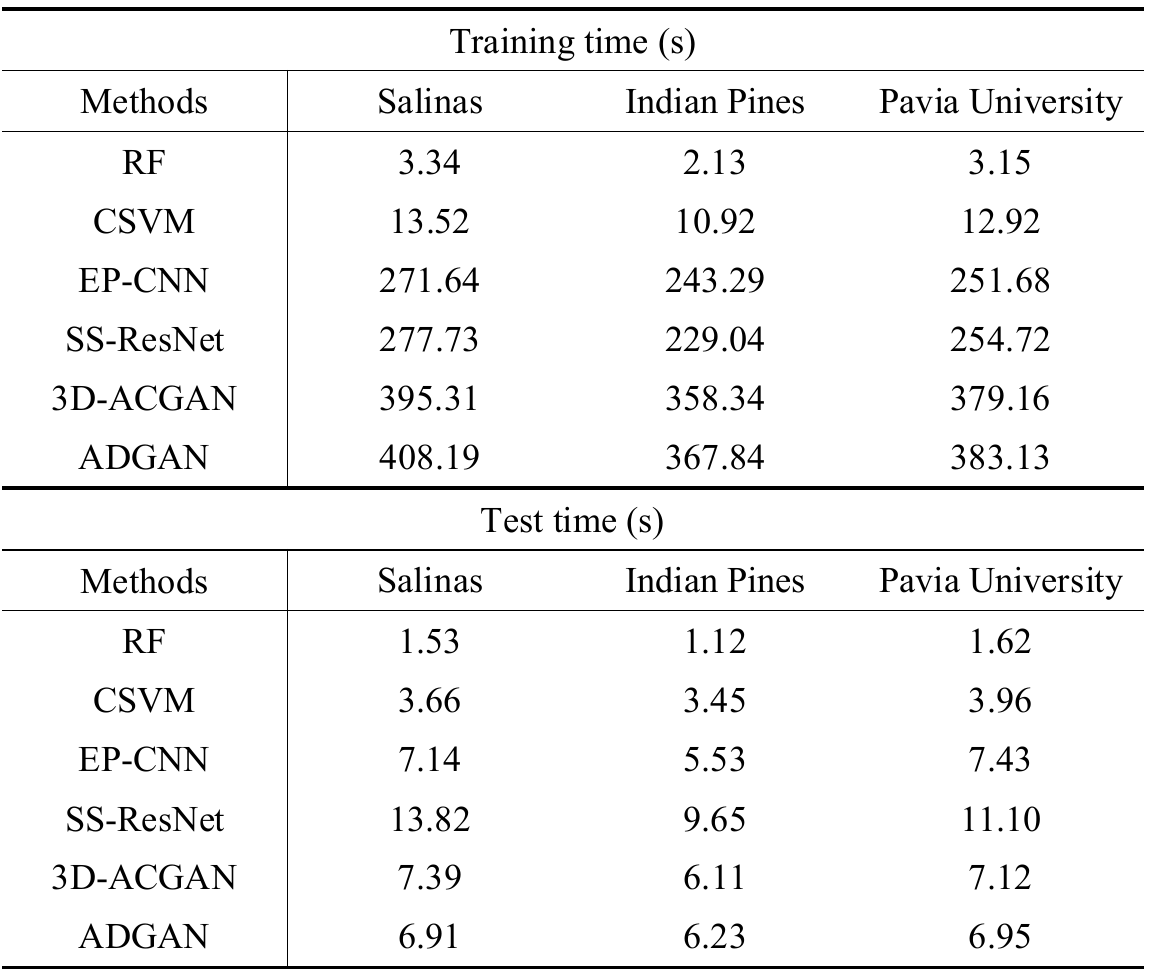}
\label{time}
\end{table}

\subsection{Visualization of Adversarial Samples}

\begin{figure}[ht]
\begin{center}
\includegraphics [width=3.4in]{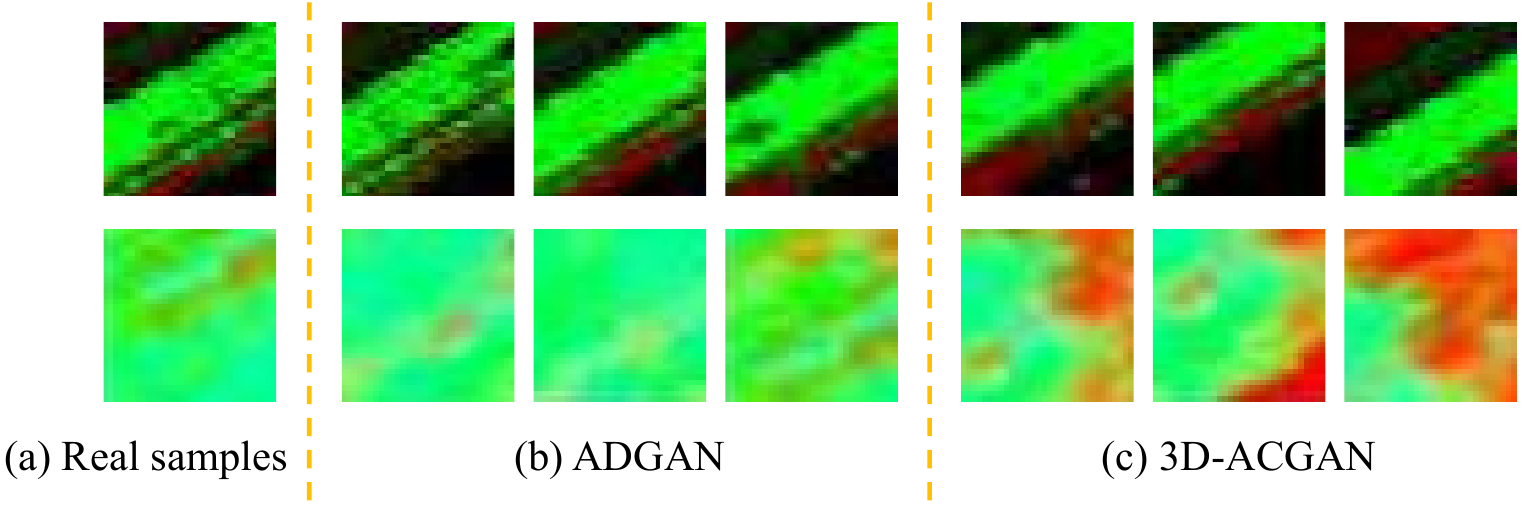}
\caption{Representative samples generated for the minority classes on the Salinas dataset.}
\label{fig_fake_img}
\end{center}
\end{figure}

In the visualization experiment, some representative fake samples generated by the 3D-ACGAN and the proposed ADGAN on the Salinas dataset are illustrated in Fig. \ref{fig_fake_img}.  As can be observed, the proposed ADGAN can generate high-quality samples which have similar structures compared with the real image samples. On the contrary, 3D-ACGAN sometimes fails to generate good samples for the minority class and collapse towards learning the basic structures of the real samples. As mentioned before, because of the self-contradiction in 3D-ACGAN's discriminator, the generator $G$ prefers to generate samples belonging to the majority class. Therefore, the performance of 3D-ACGAN in generating minority class samples is affected. The proposed ADGAN models the classification task and the discrimination task into one single objective. Hence, the mode collapse issue can be alleviated to some extent. The proposed ADGAN is superior in adversarial learning when aiming at the generation of minority class samples, and can be employed to improve the HSI classification accuracy.

\section{Conclusions and Future Work}

In this paper, an adaptive DropBlock-enhanced framework for HSI classification is proposed. The proposed ADGAN can effectively alleviate the following two problems: 1) the imbalanced training data in HSI, and 2) the mode collapse problem in GAN-based classification methods. First, the discriminator is adjusted to be a single output that returns either the fake label or the specific class label. The discriminator will not contradict itself when training samples are imbalanced. Second, AdapDrop is proposed as a regularization method to mitigate the mode collapse problem. Instead of dropping a fixed size region, the proposed AdapDrop generates drop masks with adaptive shapes, which can better deal with ground objects with various shapes. To evaluate the proposed framework, extensive experiments are performed on three hyperspectral datasets. The results show that the proposed ADGAN can achieve better performance compared with the state-of-the-art baselines.

In the future, we plan to extend our work in two directions. First, several self-attention networks will be investigated to improve classification performance. In addition, more regularization techniques will be explored to alleviate the mode collapse problem, and therefore further enhance the classification performance.

\section*{Acknowledgement}

We would like to thank Prof. Y. Chen for sharing the source code of 3D-ACGAN. We would also like to thank the Associate Editor and all the reviewers for their valuable comments and suggestions, which have significantly improved the quality of the paper.

\begin{IEEEbiography}[{\includegraphics[width=1in,height=1.25in,clip,keepaspectratio]{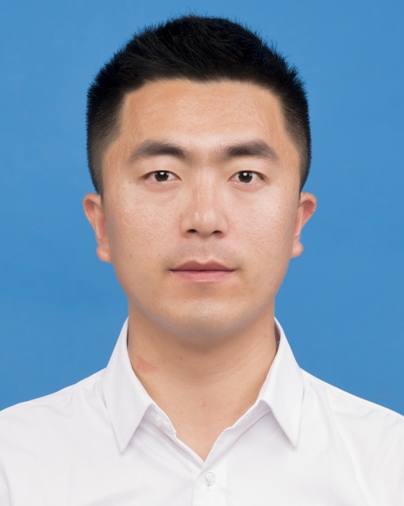}}]{Junjie Wang}
received the B. Sc. degree in computer science from Ocean University of China, Qingdao, China, in 2018. He is currently pursuing the M.Sc. degree in computer science and applied remote sensing with the School of Information Science and Technology, Ocean University of China, Qingdao, China.

His current research interests include computer vision and remote sensing image processing.

\end{IEEEbiography}

\begin{IEEEbiography}[{\includegraphics[width=1in,height=1.25in,clip,keepaspectratio]{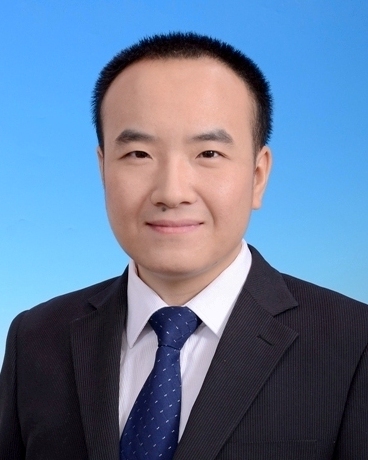}}]{Feng Gao}
received the B. Sc degree in software engineering from Chongqing University, Chongqing, China, in 2008, and the Ph. D. degree in computer science and technology from Beihang University, Beijing, China, in 2015.

He is currently an Associate Professor with the School of Information Science and Engineering, Ocean University of China. His research interests include remote sensing image analysis, pattern recognition and machine learning.

\end{IEEEbiography}

\begin{IEEEbiography}[{\includegraphics[width=1in,height=1.25in,clip,keepaspectratio]{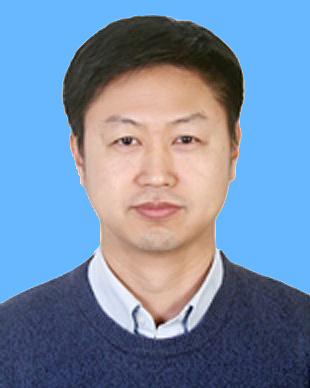}}]{Junyu Dong}
received the B.Sc. and M.Sc. degrees from the Department of Applied Mathematics, Ocean University of China, Qingdao, China, in 1993 and 1999, respectively, and the Ph.D. degree in image processing from the Department of Computer Science, Heriot-Watt University, Edinburgh, United Kingdom, in 2003.

He is currently a Professor and Vice Dean with the School of Information Science and Engineering, Ocean University of China. His research interests include visual information analysis and understanding, machine learning and underwater image processing.
\end{IEEEbiography}

\begin{IEEEbiography}[{\includegraphics[width=1in,height=1.25in,clip,keepaspectratio]{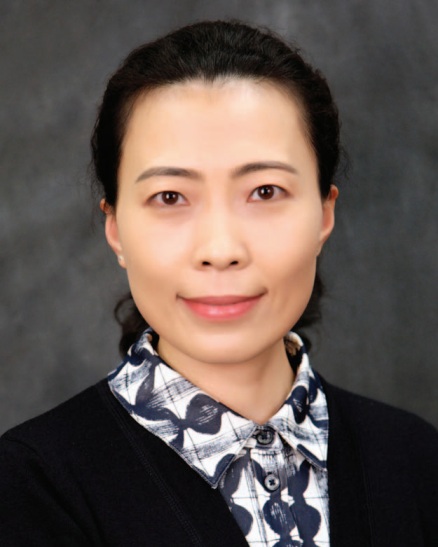}}]{Qian Du}
(Fellow, IEEE) received the Ph.D. degree in electrical engineering from the University of Maryland at Baltimore, Baltimore, MD, USA, in 2000.

She is currently the Bobby Shackouls Professor with the Department of Electrical and Computer Engineering, Mississippi State University, Starkville, MS, USA. Her research interests include hyperspectral remote sensing image analysis and applications, pattern classification, data compression, and neural networks.

Dr. Du served as a Co-Chair for the Data Fusion Technical Committee of the IEEE Geoscience and Remote Sensing Society (GRSS) from 2009 to 2013 and the Chair for the Remote Sensing and Mapping Technical Committee of the International Association for Pattern Recognition (IAPR) from 2010 to 2014. She currently serves as the Chief Editor of the \textsc{IEEE Joural of Selected Topics in Applied Earth Observations and Remote Sensing}.
\end{IEEEbiography}

\end{document}